\documentclass{article}

\PassOptionsToPackage{numbers, sort&compress}{natbib}
\usepackage[preprint]{neurips_2026}

\usepackage[utf8]{inputenc} 
\usepackage[T1]{fontenc}    
\usepackage{hyperref}       
\usepackage{url}            
\usepackage{booktabs}       
\usepackage{amsfonts}       
\usepackage{nicefrac}       
\usepackage{microtype}      
\usepackage{xcolor}         
\usepackage{graphicx}
\usepackage{amsmath}
\usepackage{amssymb}
\usepackage{multirow}
\usepackage{array}
\usepackage{mathtools}
\usepackage{overpic}
\usepackage{stmaryrd}
\usepackage{xspace}

\usepackage{my_def}

\title{Register Anything Model for Generalizable and Robust Point Cloud Registration}

\author{%
Zheng Qin$^{1}$ \hspace{5pt} Junhua Xi$^{2,3}$ \hspace{5pt} Yan Huang$^{1}$ \hspace{5pt} Jiahong Lai$^{1,2}$ \hspace{5pt} Anwen Huang$^{1}$\vspace{3pt} \\
\bf{Qiong Li$^{1}$\thanks{Corresponding authors: Qiong Li (\texttt{517066092@qq.com}), Kai Xu (\texttt{kevin.kai.xu@gmail.com}).} \hspace{5pt} Guangda Zhang$^{1}$ \hspace{5pt} Kai Xu$^{4}$\footnotemark[1]}\vspace{5pt}\\
$^{1}$\small{Defense Innovation Institute, Academy of Military Sciences, China}\\
$^{2}$\small{College of Computer Science and Technology, National University of Defense Technology, China}\\
$^{3}$\small{College of Data Intelligence, Information Support Force Engineering University, China}\\
$^{4}$\small{Institute of AI for Industry, Chinese Academy of Sciences, China}\vspace{5pt}\\
\small{\url{https://github.com/qinzheng93/RegisterAnything}}
}

\newcommand{\ours}{RAM\xspace}
\newcommand{\egtr}{EGeoTR\xspace}

\begin{document}

\maketitle

\begin{abstract}
We present Register Anything Model (\ours), a unified model for generalizable 3D point cloud registration that robustly estimates 6-DoF transformations across diverse domains spanning sensors and environments. Existing registration methods are typically tailored to a single domain, with carefully tuned architectures and hyperparameters, and thus degrade markedly when applied to novel domains. We bridge this gap by training a single model on diverse data from indoor, outdoor, and object domains. A central challenge is the severe geometric discrepancy across domains, \ie, differences in scene scale, sensor noise, point density, and structural patterns, which hinders processing with a unified point-based model. To address this, we introduce a geometry-aware normalization strategy that jointly voxelizes and normalizes point clouds into a standardized geometric space, adaptively based on their geometric properties to preserve salient local structures while standardizing resolution. Built on this normalization, \ours follows a hierarchical coarse-to-fine paradigm, and employs an efficient geometric transformer that models local geometric consistency instead of globally over all points. This design mitigates discretization noise introduced by voxelization, and improves computational efficiency. Extensive experiments on $12$ benchmarks demonstrate state-of-the-art performance in terms of inlier ratio and registration recall. Notably, our method outperforms the counterparts trained on each separate domain, showing superior generalization capability.
\end{abstract}
\section{Introduction}
\label{sec:intro}

Point cloud registration is a fundamental but critical task in 3D computer vision and robotics, which aims to estimate the $6$-D rigid transformation aligning the source point cloud to the target. It underpins numerous applications, including 3D reconstruction, LiDAR SLAM, augmented and mixed reality (AR/XR), autonomous driving and robotic manipulation.
Contemporary approaches usually follow a two-stage pipeline, which first establishes correspondences between the source and the target point clouds and then estimates the transformation based on these correspondences.

Recent learning-based methods~\cite{choy2019fully,huang2021predator,qin2022geometric,yu2023rotation,ao2023buffer,yew2020rpm} have made remarkable advances in point cloud registration. However, they are commonly tailored and trained for a specific \emph{domain}. Different domains could utilize different sensors, \eg, time-of-flight cameras \vs LiDAR scanners, or present distinct environments, \eg, indoor~\cite{zeng20173dmatch,dai2017scannet}, outdoor~\cite{geiger2012we,caesar2020nuscenes} or object~\cite{wu20153d,chang2015shapenet} environments, which induces pronounced domain shift in the geometric characteristics. 
Consequently, these methods typically succeed only in domains similar to the training domain and often fail to generalize to novel domains.
While recent work~\cite{ao2023buffer,seo2025buffer} explores adapting a model pretrained on one domain to others, target-domain geometries remain unseen during training. As a result, domain shifts are only partially mitigated. In practice, this leads to inaccurate correspondences in novel domains and degraded registration accuracy.

\begin{figure*}[t]
  \centering
  \begin{overpic}[width=0.9\linewidth]{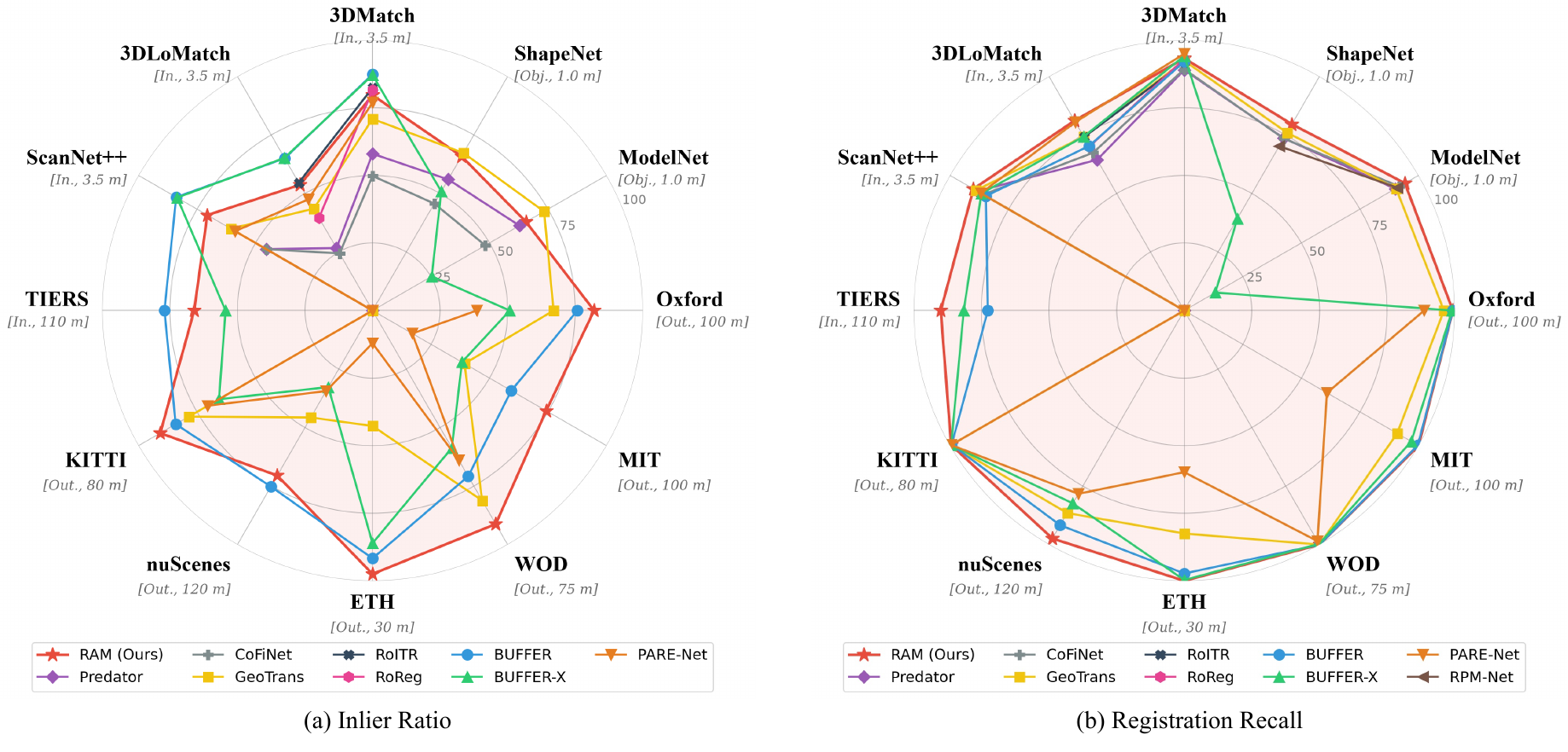}
  \end{overpic}
  \caption{Register Anything Model (\ours) achieves strong generalization and robustness across a wide range of domains. Compared with the baselines, our method extracts significantly more high-quality correspondences and effectively improves the registration performance.}
  \label{fig:teaser}
\end{figure*}

Inspired by the recent success of vision foundation models~\cite{kirillov2023segment,oquab2023dinov2,yang2024depth}, we consider the following question: \emph{\textbf{can we train a point cloud registration model with strong cross-domain generalizability by leveraging diverse point cloud data spanning a wide range of domains?}}
Our answer is \emph{\textbf{yes}}, but constructing such a unified model is challenging due to the irregular nature of point clouds and the substantial cross-domain geometric discrepancy, \eg, scene scales, point densities, and structural patterns.
At the micro-architecture level, point operators~\cite{thomas2019kpconv,choy20194d,ao2021spinnet,yao2024pare} typically encode local geometry within a fixed search radius.
Due to density variability, it is suboptimal to use the same search radius across different domains.
Dynamically adjusting search radius~\cite{seo2025buffer} requires operator kernels to model structures at diverse scales, which can complicate optimization and impair convergence.
At the macro-architecture level, current registration methods~\cite{qin2022geometric,yao2024pare} adopt domain-specific designs, \eg, downsampling factors, to better balance efficacy and efficiency: outdoor domains demand aggressive downsampling for large spatial coverage and efficiency, while indoor/object scenes benefit from finer resolutions. This heterogeneity complicates a unified architecture.

In this work, we explore cross-domain training for point cloud registration and propose a unified registration model, \emph{\textbf{R}egister \textbf{A}nything \textbf{M}odel} (\emph{\ours}). \ours follows a hierarchical coarse-to-fine paradigm~\cite{yu2021cofinet,qin2022geometric}, which first extracts coarse-grained superpoint correspondences and then refines them into fine-grained point correspondences. We address two core questions: (1) \emph{how to mitigate the geometric discrepancy across domains}, and (2) \emph{how to design an architecture that remains effective and efficient across diverse conditions}.
For the first problem, we introduce a geometry-aware normalization strategy that jointly voxelizes and normalizes point clouds into a domain-agnostic standardized geometric space, adaptively from their geometric properties. We cast this as a point cloud quality assessment problem: identify an as-large-as-possible voxel size that preserves salient local structures, \ie, does not materially alter geometry, and normalize the resulting point clouds to a reference resolution.
This allows for a unified micro architecture despite domain-dependent densities and scales.
For the second problem, we further propose an efficient geometric transformer that models inter-point-cloud geometric consistency within a limited scope rather than the whole point cloud. This design alleviates the geometric noises introduced during voxelization and effectively reduces the computational cost, contributing to a better balance between feature discriminativeness and computational efficiency.

We train \ours on a collection of $4$ registration datasets spanning indoor~\cite{zeng20173dmatch}, outdoor~\cite{geiger2012we,caesar2020nuscenes}, and object~\cite{wu20153d} benchmarks, and evaluate on $12$ benchmarks. Extensive experiments demonstrate strong generalizability and robustness across diverse domains: \ours outperforms the previous state-of-the-art BUFFER-X~\cite{seo2025buffer} in \emph{Registration Recall} on novel domains, and achieves an average improvement of $10.2$ percentage points (pp) in \emph{Inlier Ratio}. Our main contributions are:
\begin{itemize}[leftmargin=2em,itemsep=0pt,topsep=0pt]
\item We study the efficacy of cross-domain training for point cloud registration and propose a unified registration model with strong cross-domain generalizability and robustness.
\item We introduce a geometry-aware normalization strategy which jointly voxelizes and normalizes point clouds into a regularized geometry space, mitigating cross-domain geometric discrepancy.
\item We devise an efficient geometric transformer which models inter-point-cloud geometric consistency within a limited scope, achieving a better balance between efficiency and efficacy.
\item We conduct extensive experiments across a wide range of domains and show that our method establishes new state-of-the-art performance on $12$ benchmarks.
\end{itemize}


\section{Related Work}
\label{sec:related}

\subsection{Point Cloud Registration}

Point cloud registration plays a critical role in various computer vision and robotics applications. Most registration methods follow a matching-based pipeline.
First, a set of correspondences between the source and the target point clouds are established by geometric feature matching. Recently, learning-based methods~\cite{choy2019fully,huang2021predator,yu2021cofinet,ao2021spinnet,qin2022geometric,wang2022you,yu2023rotation,ao2023buffer,yao2024pare,huang2022imfnet,huang2025psreg} have made remarkable advances by learning discriminative geometric features in a data-driven manner. Early methods~\cite{choy2019fully,bai2020d3feat,huang2021predator} usually first detect a set of keypoints and then match them based on their features. However, as keypoint detection is an ill-posed problem, it suffers from low repeatability and thus harms the registration performance. Recent methods~\cite{yu2021cofinet,qin2022geometric,yu2023rotation,ao2023buffer,yao2024pare} bypass keypoint detection by employing a hierarchical registration framework, which first extracts coarse-grained superpoint or patch correspondences and then refine them to fine-grained point correspondences.
And some methods~\cite{ao2021spinnet,qin2022geometric,wang2022you,ao2023buffer,yao2024pare} design rotation-invariant or rotation-equivariant operators to improve the robustness to large rotations.
After correspondences are extracted, the rigid transformation can be estimated by robust estimators such as RANSAC~\cite{fischler1981random,chen2022sc2,zhang20233d} and learning-based estimators~\cite{choy2020deep,bai2021pointdsc}.
There are also end-to-end methods which directly regress and optimize the transformation~\cite{aoki2019pointnetlk,yew2020rpm,fu2021robust,xu2021omnet}. However, they are vulnerable to complex geometric structures and severe occlusion, which limits their application in real-world applications.

\subsection{Generalizable Point Cloud Registration}

Although recent learning-based methods have made remarkable progress, they usually suffer from weak generalizability across different domains due to their significant geometric discrepancy, such as scene scales, point density, noise distributions and structural patterns.
A model trained on indoor datasets usually has difficulty in registering outdoor point clouds, and vice versa.
Recent methods~\cite{ao2023buffer,seo2025buffer} have studied to adapt a registration model pretrained on one domain to more domains.
BUFFER-X~\cite{seo2025buffer} adaptively selects the voxel size based on the sphericity of point clouds and the search radius by considering the point density, which generalizes the BUFFER~\cite{ao2023buffer} model trained on 3DMatch to outdoor scenes.
However, due to the lack of awareness of domain-specific geometries during training, it still suffers from inaccurate correspondences in novel domains, which leads to suboptimal registration in hard cases.
Besides, it still needs specific tuning for different domains, \eg, inlier thresholds for outlier rejection and RANSAC hyperparameters, which limits its application in real-world scenarios.
In this work, we bridge this gap by constructing a unified registration model by cross-domain training, which achieves strong generalizability and robustness.

\subsection{Cross-source Point Cloud Registration}

Cross-source registration~\cite{huang2017systematic,huang2017coarse,huang2023cross,huang2103comprehensive,zhao2024vrhcf,zhao2025cross} is another closely related but distinct problem to cross-domain registration.
Both must handle substantial variations in point density, geometric scale, sensor noise, and sampling patterns.
Cross-source registration aims to align two point clouds acquired from different sources or sensing modalities, such as Kinect, LiDAR, and structure-from-motion systems.
In contrast, cross-domain registration aims to develop a unified model that can register point-cloud pairs across different environments or sensors.
In this work, the two point clouds within each pair are generally acquired from the same source, while the source or environment may vary across pairs.
We discuss the effectiveness of our method in cross-source registration in the supplementary material.
\section{Method}
\label{sec:method}

\subsection{Overview}
\label{sec:method:overview}

Given two point clouds $\mathcal{P} = \{ \mathbf{p}_i \in \mathbb{R}^3 \mid i = 1, \dots, M \}$ and $\mathcal{Q} = \{ \mathbf{q}_j \in \mathbb{R}^3 \mid j = 1, \dots, N \}$, our goal is to estimate the rigid transformation $\mathbf{T} \in \SE(3)$ aligning $\mathcal{P}$ to $\mathcal{Q}$, where $\mathbf{T}$ consists of a 3D rotation $\mathbf{R} \in \SO(3)$ and a 3D translation $\mathbf{t} \in \mathbb{R}^3$. Our method follows the traditional registration pipeline where we first extract point correspondences $\mathcal{C} = \{ (\mathbf{x}_i, \mathbf{y}_i) \mid \mathbf{x}_i \in \mathcal{P}, \mathbf{y}_i \in \mathcal{Q}\}$, and then estimate $\mathbf{T}$ based on the correspondences by solving:
\begin{equation}
\mathbf{R}, \mathbf{t} = \argmin_{\mathbf{R} \in \SO(3), \mathbf{t} \in \mathbb{R}^3} \sum_{(\mathbf{x}_i, \mathbf{y}_i) \in \mathcal{C}} \lVert \mathbf{R} \mathbf{x}_i + \mathbf{t} - \mathbf{y}_i \rVert^ 2.
\end{equation}

\begin{figure*}[t]
  \centering
  \begin{overpic}[width=\linewidth]{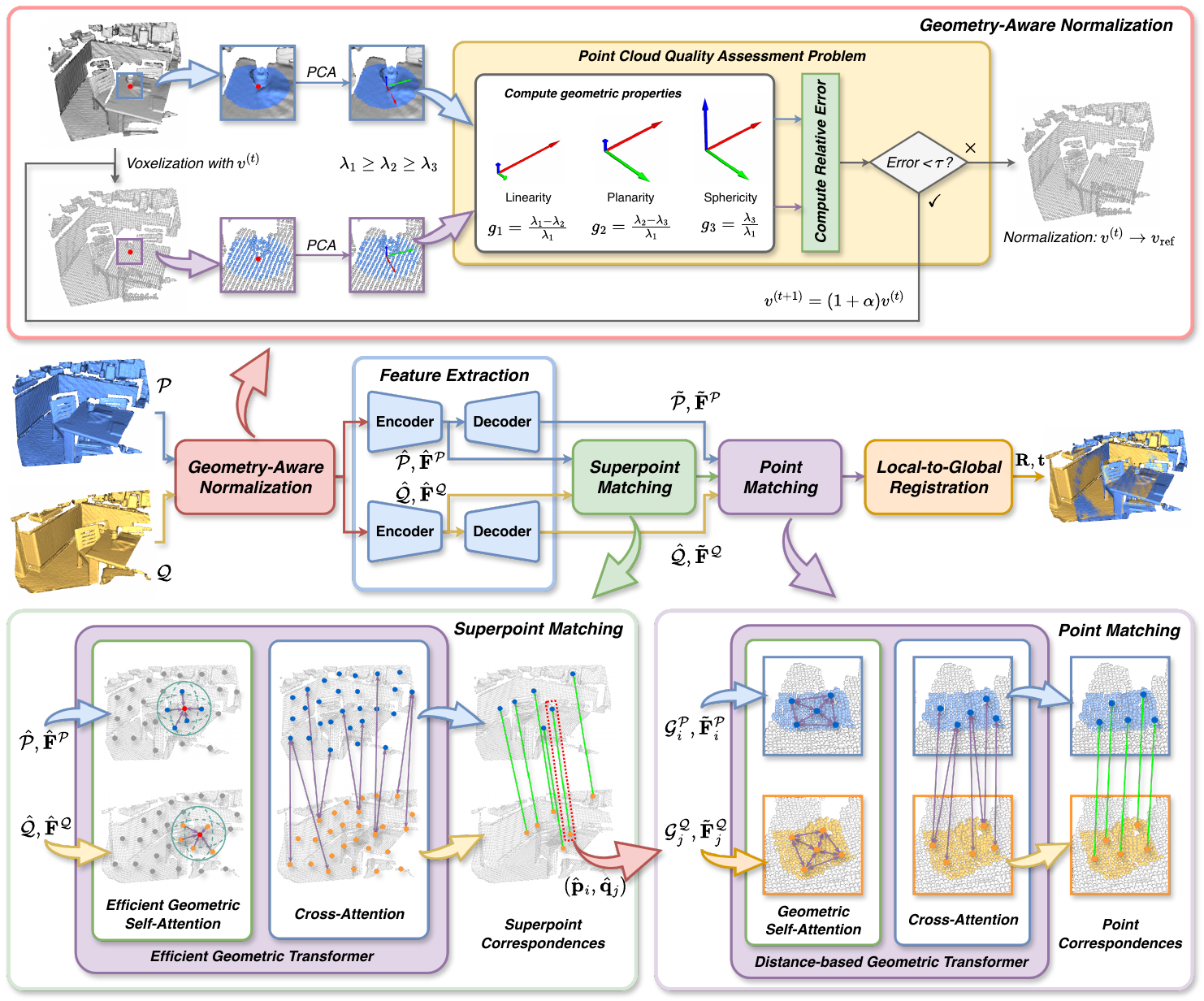}
  \end{overpic}
  \caption{Overall pipeline of \ours. Point clouds are first adaptively voxelized and normalized into a standardized geometric space by a geometry-aware normalization module. Aftering extracting hierarchical features, \ours extracts superpoints correspondences with an \egtr, and then refines them into dense point correspondences. At last, the transformation is recovered from the correspondences.}
  \label{fig:pipeline}
\end{figure*}

Previous methods~\cite{huang2021predator,yu2021cofinet,qin2022geometric,yao2024pare,ao2023buffer} assume that point clouds to register share the training domain, leading to poor generalizability across domains.
In this work, we break this gap by jointly training a unified point cloud registration model on data from various domains.
To address cross-domain geometric discrepancy, we first design a geometry-aware normalization module which jointly voxelizes and normalizes point clouds into a domain-agnostic standardized geometric space (\cref{sec:method:gsn}).
\ours then employs a hierarchical registration pipeline~\cite{yu2021cofinet,qin2022geometric,yao2024pare}, which first extracts superpoint correspondences and then refine them into dense point correspondences (\cref{sec:method:ram}).
Finally, the rigid transformation is estimated by LGR~\cite{qin2022geometric}. \cref{fig:pipeline} illustrates the overall pipeline.

\subsection{Geometry-aware Normalization}
\label{sec:method:gsn}

A unified registration model must handle the geometric discrepancy among point clouds from different domains, \eg, in scene scale and point density. For instance, outdoor LiDAR scans are sparse and long-range (typically ${\sim}100\text{m}$), while indoor point clouds are denser and short-range (typically ${<}10\text{m}$). Modern point operators~\cite{thomas2019kpconv,xu2021paconv,yao2024pare,ao2021spinnet,choy20194d,yu2023rotation,ao2023buffer,choy2019fully} encode local geometry within a fixed search radius, which is problematic across domains: a small radius on large-scale scenes yields insufficient receptive fields, whereas a large radius on small-scale scenes oversmooths local details.

To address this issue, we design a \emph{geometry-aware normalization} module to jointly voxelize and normalize the point clouds into a domain-agnostic standardized geometric space with consistent scales and densities, enabling effective processing with a unified search radius.
Our key motivation is to adaptively select a voxel size $v_{\text{raw}}$ per point cloud pair according to their geometric properties, and then align it with a reference $v_{\text{ref}}$ by rescaling $\mathbf{p}_{\text{norm}} = \nicefrac{v_{\text{ref}}}{v_{\text{raw}}} \cdot \mathbf{p}$.
We formulate the selection of $v_{\text{raw}}$ as a \emph{point cloud quality assessment problem}, and iteratively seek for an as-large-as-possible voxel size whose induced geometric change remains tolerable.

Given raw point clouds $\mathcal{P}$ and $\mathcal{Q}$, we initialize $v^{(0)} = \phi(\min(d(\mathcal{P}), d(\mathcal{Q})))$, where $d(\cdot)$ is the diagonal length of the bounding box, and $\phi(\cdot)$ is a monotonically increasing function (see the appendix for the definition).
In $t$-th iteration, we voxelize $\mathcal{P}$ with $v^{(t)}$ to obtain $\mathcal{P}'$, and sample a set of seeds $\{ \mathbf{s}_i \}$ from $\mathcal{P}'$ via furthest point sampling.
For each seed $\mathbf{s}_i$, we collect the local patches $\mathcal{N}_i$ and $\mathcal{N}'_i$ within the radius of $r = \lambda \cdot v^{(t)}$ in $\mathcal{P}$ and $\mathcal{P}'$, respectively, where $\lambda$ is a scaling factor.
Inspired by~\cite{alexiou2021pointpca}, we compute the eigenvalues $\lambda_1 \ge \lambda_2 \ge \lambda_3$ of the covariance matrix of each patch and derive three geometric properties, \ie, linearity $g_1$, planarity $g_2$ and sphericity $g_3$:
\begin{equation}
g_1 = (\lambda_1 - \lambda_2) / \lambda_1,\hspace{8pt} g_2 = (\lambda_2 - \lambda_3) / \lambda_1,\hspace{8pt} g_3 = \lambda_3 / \lambda_1.
\end{equation}
We define the geometric error $e_i$ of $\mathbf{s}_i$ by the maximum relative error on linearity and planarity between $\mathcal{N}_i$ and $\mathcal{N}'_i$.
The geometric error $e(\mathcal{P})$ between $\mathcal{P}$ and $\mathcal{P}'$ is the median geometric error over the half of seeds with the largest sphericity in $\mathcal{P}$. The same computation goes for $\mathcal{Q}$.
Our insight behind this design is that oversmoothing primarily alters linearity and planarity, while sphericity serves as an indicator of local geometric complexity, filtering out geometry-less patches which are less sensitive to oversmoothing.
If $e(\mathcal{P})$ and $e(\mathcal{Q})$ are both below a tolerance $\tau$, we enlarge the voxel size by $v^{(t+1)} = (1 + \alpha) v^{(t)}$ and repeat the above process until $e \ge \tau$ or by up to $T$ times.

\subsection{Register Anything Model}
\label{sec:method:ram}

Given normalized point clouds, \ours performs coarse-to-fine registration~\cite{yu2021cofinet,qin2022geometric,yao2024pare}, \ie, superpoint matching followed by dense point matching. For brevity, we omit the prime and reuse $\mathcal{P}$ and $\mathcal{Q}$ for the normalized point clouds in the following text.

\paragraph{Feature extraction.}
We extract hierarchical point features with a KPConv-FPN~\cite{thomas2019kpconv} backbone.
Following~\cite{qin2022geometric}, the coarsest-level points $\hat{\mathcal{P}}$ and $\hat{\mathcal{Q}}$ are used as the superpoints, while the half-resolution points $\tilde{\mathcal{P}}$ and $\tilde{\mathcal{Q}}$ support dense point matching.
Their features are denoted as $\hat{\mathbf{F}}^{\mathcal{P}} \in \mathbb{R}^{\hat{M} \times d_{\text{c}}}$, $\hat{\mathbf{F}}^{\mathcal{Q}} \in \mathbb{R}^{\hat{N} \times d_{\text{c}}}$, $\tilde{\mathbf{F}}^{\mathcal{P}} \in \mathbb{R}^{\tilde{M} \times d_{\text{f}}}$ and $\tilde{\mathbf{F}}^{\mathcal{Q}} \in \mathbb{R}^{\tilde{N} \times d_{\text{f}}}$.
The dense points are assigned to their nearest superpoints via point-to-node partition.
The points assigned to $\hat{\textbf{p}}_i$ and $\hat{\textbf{q}}_j$ are denoted as $\mathcal{G}^{\mathcal{P}}_i$ and $\mathcal{G}^{\mathcal{Q}}_j$, with gathered features $\tilde{\mathbf{F}}^{\mathcal{P}}_i \subset \tilde{\mathbf{F}}^{\mathcal{P}}$ and $\tilde{\mathbf{F}}^{\mathcal{Q}}_j \subset \tilde{\mathbf{F}}^{\mathcal{Q}}$.

\para{Superpoint matching.}
Geometric transformer~\cite{qin2022geometric} improves superpoint feature discriminativeness by encoding inter-point-cloud global geometric consistency~\cite{qin2022geometric,yu2023rotation,yu2023peal,yao2024pare}, but their geometric structure embedding has memory complexity $O(N^2 d)$, which is prohibitive for large point clouds and typically necessitates domain-specific downsampling~\cite{qin2022geometric,yao2024pare,yu2023rotation}.
Moreover, voxelization can destabilize the geometric consistency between distant superpoints, and we observe that attention scores for such pairs are empirically low, indicating redundancy in the global geometric structure embedding.

We therefore propose an \emph{efficient geometric transformer} (\emph{\egtr}) module, which models the local geometric consistency between point clouds instead.
\egtr interleaves local geometric self-attention encoding intra-point-cloud local geometric structures and cross-attention encoding inter-point-cloud geometric consistency to produce reliable superpoint features.
Specifically, for each superpoint $\hat{\textbf{p}}_i \in \hat{\mathcal{P}}$, let $\mathcal{N}_i$ be the indices of its $k$-nearest superpoints. Given input features $\textbf{X} \in \mathbb{R}^{\hat{M} \times d_{\text{c}}}$, the local geometric self-attention produces output features $\textbf{Z} \in \mathbb{R}^{\hat{M} \times d_{\text{c}}}$ as:
\begin{equation}
\mathbf{z}_i = \sum_{j \in \mathcal{N}_i} \frac{\exp(a_{i, j})}{\sum_{k \in \mathcal{N}_i} \exp(a_{i, k})} (\mathbf{x}_j \mathbf{W}^{\text{V}}),
\end{equation}
and the attention score $a_{i, j}$ is computed as:
\begin{equation}
a_{i, j} = \frac{(\mathbf{x}_i \mathbf{W}^{\text{Q}}) (\mathbf{x}_j \mathbf{W}^{\text{K}} + \mathbf{r}_{i, j} \mathbf{W}^{\text{R}})^{\top}}{\sqrt{d_{\text{c}}}},
\end{equation}
where $\mathbf{r}_{i, j}$ is the geometric structure embedding~\cite{qin2022geometric}, combining pair-wise distance embedding and triplet-wise angular embedding. $\mathbf{W}^{\text{Q}}, \mathbf{W}^{\text{K}}, \mathbf{W}^{\text{V}}, \mathbf{W}^{\text{R}} \in \mathbb{R}^{d_{\text{c}} \times d_{\text{c}}}$ are the learnable projection weights for queries, keys, values and embeddings, respectively.
The same computation goes for $\hat{\mathcal{Q}}$.
This design reduces the memory complexity of the geometric structure embedding to $O(Nkd)$, where $k \ll N$, and therefore avoids the need to alter downsampling ratios for various domains.
For the cross-attention, since the geometric structure embedding is not injected, it can be easily accelerated by flash attention~\cite{shah2024flashattention} or replaced by linear attention~\cite{wang2020linformer}.
The resulting hybrid superpoint features are denoted as $\hat{\mathbf{H}}^{\mathcal{P}}$ and $\hat{\mathbf{H}}^{\mathcal{Q}}$.

At last, the superpoint correspondences are extracted by selecting the largest $N_c$ entries in the correlation matrix between $\hat{\mathbf{H}}^{\mathcal{P}}$ and $\hat{\mathbf{H}}^{\mathcal{Q}}$. Following~\cite{qin2022geometric}, we adopt dual normalization on the correlation matrix to suppress the ambiguous matches.

\para{Point matching.}
Given a superpoint correspondence $(\hat{\mathbf{p}}_{i}, \hat{\mathbf{q}}_{j})$, we extract dense point correspondences by matching the points in $\mathcal{G}^{\mathcal{P}}_{i}$ and $\mathcal{G}^{\mathcal{Q}}_{j}$.
Previous work~\cite{qin2022geometric,yu2023rotation,yang2022one,yao2024pare} directly match the geometric features $\tilde{\mathbf{F}}^{\mathcal{P}}_{i}$ and $\tilde{\mathbf{F}}^{\mathcal{Q}}_{j}$ from the backbone. However, within each patch, the points are so close that they have very similar features and cannot be effectively distinguished. 
We therefore employ a lightweight \egtr module comprising one local geometric self-attention and one cross-attention.
Unlike the superpoint matching, we only use the pair-wise distance embedding in the local geometric self-attention for better efficiency.
The resulting hybrid point features are denoted as $\tilde{\mathbf{H}}^{\mathcal{P}}_{i}$ and $\tilde{\mathbf{H}}^{\mathcal{Q}}_{j}$.
Inspired by~\cite{lindenberger2023lightglue}, we further predict the overlapping scores $\tilde{\mathbf{o}}^{\mathcal{P}}_{i}$ and $\tilde{\mathbf{o}}^{\mathcal{Q}}_{j}$ of the dense points and refines the point features as $\tilde{\mathbf{Y}}^{\mathcal{P}}_{i}$ and $\tilde{\mathbf{Y}}^{\mathcal{Q}}_{j}$ with two respective linear projections:
\begin{equation}
\begin{aligned}
\tilde{\mathbf{o}}^{\mathcal{P}}_{i} = \sigma(\tilde{\mathbf{H}}^{\mathcal{P}}_{i} \mathbf{w}^{\text{O}}), \hspace{5pt} & \tilde{\mathbf{o}}^{\mathcal{Q}}_{j} = \sigma(\tilde{\mathbf{H}}^{\mathcal{Q}}_{j} \mathbf{w}^{\text{O}}), \\
\tilde{\mathbf{Y}}^{\mathcal{P}}_{i} = \tilde{\mathbf{H}}^{\mathcal{P}}_{i} \mathbf{W}^{\text{F}}, \hspace{5pt} & \tilde{\mathbf{Y}}^{\mathcal{Q}}_{j} = \tilde{\mathbf{H}}^{\mathcal{Q}}_{j} \mathbf{W}^{\text{F}}, 
\end{aligned}
\end{equation}
where $\mathbf{w}^{\text{O}} \in \mathbb{R}^{d_{\text{f}}}$ and $\mathbf{W}^{\text{F}} \in \mathbb{R}^{d_{\text{f}} \times d_{\text{f}}}$ are learnable weights, and $\sigma(\cdot)$ is the sigmoid function.

The point correspondences are extracted by jointly considering the overlapping scores and the feature similarity. 
Specifically, we compute the similarity matrix $\tilde{\mathbf{S}} = \tilde{\mathbf{Y}}^{\mathcal{P}}_{i} (\tilde{\mathbf{Y}}^{\mathcal{Q}}_{j})^{\top}$ between $\mathcal{G}^{\mathcal{P}}_{i}$ and $\mathcal{G}^{\mathcal{Q}}_{j}$, and inject the overlapping scores to obtain the confidence matrix $\mathbf{S}$:
\begin{equation}
\mathbf{S} = \left[ \tilde{\mathbf{o}}^{\mathcal{P}}_{i} (\tilde{\mathbf{o}}^{\mathcal{Q}}_{j})^{\top} \right] \odot \varphi_{\text{row}}(\tilde{\mathbf{S}}) \odot \varphi_{\text{col}}(\tilde{\mathbf{S}}),
\end{equation}
where $\odot$ denotes the Hadamard product and $\varphi_{\text{row}}(\cdot)$ and $\varphi_{\text{col}}(\cdot)$ are the row-wise and column-wise softmax operations.
The local point correspondences are extracted by mutual top-$k$ selection on $\mathbf{S}$, and the global point correspondences aggregate the local correspondences from all superpoint correspondences.
The final transformation is estimated by LGR~\cite{qin2022geometric} from the point correspondences.

\subsection{Loss Functions}

\ours is trained with three loss functions: (1) an overlap-aware circle loss~\cite{qin2022geometric} $\mathcal{L}_{\text{oc}}$ to supervise the superpoint features, (2) a negative log-likelihood loss $\mathcal{L}_{\text{nll}}$ to supervise the confidence matrices, and (3) a binary cross-entropy loss $\mathcal{L}_{\text{bce}}$ to supervise the overlapping scores. The overall loss is $\mathcal{L} = \lambda_{\text{oc}} \mathcal{L}_{\text{oc}} + \lambda_{\text{nll}} \mathcal{L}_{\text{nll}} + \lambda_{\text{bce}} \mathcal{L}_{\text{bce}}$.
\section{Experiments}
\label{sec:exp}

\subsection{Benchmarks and metrics.}
\label{sec:exp:impl}

\para{Benchmarks.}
Following~\cite{seo2025buffer}, we evaluate \ours on $12$ point cloud registration benchmarks, including $4$ indoor benchmarks: 3DMatch~\cite{zeng20173dmatch}, 3DLoMatch~\cite{huang2021predator}, ScanNet++~\cite{yeshwanth2023scannet++} and TIERS~\cite{li2022multi}, $6$ outdoor benchmarks: KITTI~\cite{geiger2012we}, nuScenes~\cite{caesar2020nuscenes}, ETH~\cite{pomerleau2012challenging}, MIT~\cite{tian2023resilient}, WOD~\cite{sun2020scalability,liu2024extend} and Oxford~\cite{ramezani2020newer}, and $2$ object benchmarks: ModelNet~\cite{wu20153d} and ShapeNet~\cite{chang2015shapenet}.
We train \ours on a hybrid set combining 3DMatch, KITTI, nuScenes and ModelNet. Additional details are in the appendix.

\para{Metrics.}
Following~\cite{qin2022geometric,seo2025buffer}, we report the results on $4$ metrics:
(1) \emph{Inlier Ratio} (IR), the percentage of correspondences whose residuals are below a certain threshold under the ground-truth transformation.
(2) \emph{Relative Rotation Error} (RRE), the geodesic distance between the estimated and the ground-truth rotations.
(3) \emph{Relative Translation Error} (RTE), the $\ell_2$-distance between the estimated and the ground-truth translations.
(4) \emph{Registration Recall} (RR), the percentage of successfully registered pairs. A pair is successful if its RMSE is below $0.2$m (3DMatch/3DLoMatch) or if both RRE and RTE are below certain thresholds (other benchmarks). 
We adopt stricter RRE and RTE thresholds to enable more rigorous performance assessment.
Please refer to the appendix for more details.

\subsection{Experiments on Indoor Benchmarks}
\label{sec:exp:indoor}


\begin{table}[!t]
\setlength{\tabcolsep}{8pt}
\scriptsize
\centering
\caption{
Results on indoor benchmarks.
\textbf{Boldfaced} numbers highlight the best and the second best are \underline{underlined}.
$^{\dagger}$ and $^{\ddagger}$ denote the models pretrained on 3DMatch and KITTI, respectively.
$^{\triangledown}$ denotes the models without outlier rejection.
$^{\star}$ denotes the model trained on indoor data.
``N/A'' indicates the model fails to achieve reasonable registrations.
}
\label{table:results-indoor}
\begin{tabular}{lcccccccc}
\toprule
\multirow{2}{*}{Model} & \multicolumn{2}{c}{3DMatch} & \multicolumn{2}{c}{3DLoMatch} & \multicolumn{2}{c}{ScanNet++} & \multicolumn{2}{c}{TIERS} \\
\cmidrule(r){2-9}
 & IR(\%) & RR(\%) & IR(\%) & RR(\%) & IR(\%) & RR(\%) & IR(\%) & RR(\%) \\
\midrule
Predator~\cite{huang2021predator} & 58.0 & 89.0 & 26.7 & 64.4 & 45.4 & 88.3 & \multicolumn{2}{c}{N/A} \\
CoFiNet~\cite{yu2021cofinet} & 49.8 & 89.3 & 24.4 & 67.4 & 45.4 & 85.8 & \multicolumn{2}{c}{N/A} \\
GeoTrans~\cite{qin2022geometric} & 70.9 & 92.4 & 43.5 & 74.3 & 60.6 & \underline{88.9} & \multicolumn{2}{c}{N/A} \\
RoITR~\cite{yu2023rotation} & 82.4 & 91.9 & 54.6 & 74.1 & - & - & - & - \\
BUFFER~\cite{ao2023buffer} & \textbf{87.4} & 92.1 & \underline{65.0} & 70.3 & \textbf{83.9} & 84.7 & \textbf{77.0} & 72.6 \\
RoReg~\cite{wang2023roreg} & 81.6 & 93.0 & 39.6 & 70.1 & - & - & - & - \\
PARENet~\cite{yao2024pare} & 76.9 & \textbf{95.0} & 47.5 & \underline{80.5} & 58.7 & 86.8 & \multicolumn{2}{c}{N/A} \\
\midrule
BUFFER-X$^{\dagger}$~\cite{seo2025buffer} & \underline{87.1} & \underline{94.2} & \textbf{65.2} & 74.3 & \underline{83.5} & 86.6 & 54.4 & 81.5 \\
BUFFER-X$^{\ddagger}$~\cite{seo2025buffer} & 83.1 & 92.3 & 59.5 & 67.1 & 79.2 & 86.6 & 60.8 & \underline{84.6} \\
\midrule
BUFFER$^{\triangledown}$~\cite{ao2023buffer} & 50.2 & 89.6 & 21.1 & 60.9 & 36.5 & 64.5 & 44.1 & 69.1 \\
BUFFER-X$^{\triangledown\dagger}$~\cite{seo2025buffer} & 35.1 & 90.2 & 14.0 & 56.7 & 26.7 & 83.3 & 31.7 & 81.9 \\
\midrule
\ours{}$^{\star}$ (\emph{ours}) & 75.0 & 92.4 & 47.7 & 77.8 & 65.9 & 89.2 & 48.5 & 81.3 \\
\ours (\emph{ours}) & 79.7 & 93.3 & 53.7 & \textbf{81.1} & 70.6 & \textbf{90.0} & \underline{66.0} & \textbf{90.0} \\
\bottomrule
\end{tabular}
\end{table}

\para{Baselines.}
We compare against $9$ baselines: Predator~\cite{huang2021predator}, CoFiNet~\cite{yu2021cofinet}, GeoTrans~\cite{qin2022geometric}, RoITR~\cite{yu2023rotation}, BUFFER~\cite{ao2023buffer}, RoReg~\cite{wang2023roreg}, PEAL~\cite{yu2023peal}, PARENet~\cite{yao2024pare} and BUFFER-X~\cite{seo2025buffer}. The first $8$ baselines are trained on 3DMatch. And we compare with two BUFFER-X variants based on BUFFER pretrained on 3DMatch (marked by $^{\dagger}$) and KITTI (marked by $^{\ddagger}$), respectively.

\para{Quantitative results on 3DMatch and 3DLoMatch.}
\cref{table:results-indoor}~(left) shows the results on 3DMatch and 3DLoMatch.
For \emph{Inlier Ratio}, \ours is better than or on par with most baselines except BUFFER. This is because BUFFER integrates a learnable outlier rejection module, which significantly improves the inlier ratio. 
For \emph{Registration Recall}, \ours outperforms all baselines on the more challenging 3DLoMatch, demonstrating the strong robustness of our method to severe occlusion.
And \ours trained on hybrid data also surpasses the model trained on in-domain data (marked by $^{\star}$).
These results indicate that exposure to diverse domains enriches geometric patterns during training, which effectively enhances the learned point features and yields more accurate correspondences. This is consistent with our goal to construct a unified registration model via cross-domain training.

\para{Quantitative results on unseen benchmarks.}
We further evaluate the generalizability on two unseen indoor benchmarks, ScanNet++ and TIERS.
\ours surpasses BUFFER-X respectively by $3.4$ and $5.4$ percentage points (pp) in \emph{Registration Recall} on ScanNet++ and TIERS, respectively, and by $5.2$ pp in \emph{Inlier Ratio} on TIERS.
These results have demonstrated the superior generalizability of our method to unseen domains.
Note that ScanNet++ has a similar data distribution with 3DMatch, so BUFFER-X based on 3DMatch outperforms the variant based on KITTI. While TIERS is captured with a spinning LiDAR scanner and thus resembles an outdoor domain, favouring the KITTI-based variant.
On the contrary, our method consistently outperforms both variants of BUFFER-X, further proving the efficacy of cross-domain training for improving the registration generalizability.

\para{Comparison with BUFFER and BUFFER-X without outlier rejection.}
We additionally report the performance of BUFFER and BUFFER-X before the outlier-rejection module (marked by $^{\triangledown}$).
The dedicated outlier-rejection stage substantially improves their performance.
Before outlier rejection, \ours consistenctly outperforms both baselines across all benchmarks, indicating that our method produces more robust initial correspondences.
After outlier rejection, \ours still achieves higher \emph{Registration Recall} on 3DLoMatch, ScanNet++, and TIERS, demonstrating stronger generalization to challenging unseen indoor benchmarks.

\subsection{Experiments on Outdoor Benchmarks}
\label{sec:exp:outdoor}


\begin{table}[!t]
\setlength{\tabcolsep}{4.5pt}
\scriptsize
\centering
\caption{
Results on outdoor benchmarks.
\textbf{Boldfaced} numbers highlight the best and the second best are \underline{underlined}.
$^{\dagger}$ and $^{\ddagger}$ denote the models pretrained on 3DMatch and KITTI, respectively.
$^{\star}$ denotes the model trained on outdoor data.
$^{\diamond}$ denotes the model trained without nuScenes.
}
\label{table:results-outdoor}
\begin{tabular}{lcccccccccccc}
\toprule
\multirow{2}{*}{Model} & \multicolumn{2}{c}{KITTI} & \multicolumn{2}{c}{nuScenes} & \multicolumn{2}{c}{ETH} & \multicolumn{2}{c}{WOD} & \multicolumn{2}{c}{MIT} & \multicolumn{2}{c}{Oxford} \\
\cmidrule(r){2-13}
 & IR(\%) & RR(\%) & IR(\%) & RR(\%) & IR(\%) & RR(\%) & IR(\%) & RR(\%) & IR(\%) & RR(\%) & IR(\%) & RR(\%) \\
\midrule
GeoTrans~\cite{qin2022geometric} & 78.5 & \textbf{99.5} & 45.7 & 86.5 & 42.7 & 82.5 & 81.2 & \textbf{100.0} & 39.3 & 90.9 & 67.0 & 96.0 \\
BUFFER~\cite{ao2023buffer} & 84.1 & 99.3 & \textbf{75.2} & 91.7 & 91.7 & 97.3 & 70.8 & \textbf{100.0} & 59.3 & \underline{99.7} & 75.8 & \textbf{99.3} \\
PARENet~\cite{yao2024pare} & 70.5 & \textbf{99.5} & 34.4 & 78.3 & 12.1 & 59.8 & 64.0 & 98.5 & 17.0 & 60.9 & 38.6 & 88.7 \\
\midrule
BUFFER-X$^{\dagger}$~\cite{seo2025buffer} & 65.6 & \textbf{99.5} & 32.7 & 82.4 & 86.1 & 99.7 & 58.7 & 99.7 & 38.2 & 96.9 & 50.8 & 99.0 \\
BUFFER-X$^{\ddagger}$~\cite{seo2025buffer} & 73.9 & \textbf{99.5} & 52.6 & 93.5 & 90.8 & \textbf{100.0} & 65.2 & 99.7 & 51.6 & 98.7 & 59.4 & \textbf{99.3} \\
\midrule
\ours{}$^{\star}$ (\emph{ours}) & 89.5 & \textbf{99.5} & 70.0 & \underline{96.8} & \underline{93.8} & \underline{97.9} & \underline{90.4} & \textbf{100.0} & \underline{72.3} & \underline{98.7} & \underline{79.2} & \underline{99.0} \\
\ours{}$^{\diamond}$ (\emph{ours}) & \underline{90.5} & \textbf{99.5} & 59.4 & 95.2 & \underline{96.7} & 99.3 & \underline{90.0} & \textbf{100.0} & \underline{62.5} & 97.4 & \underline{80.6} & \textbf{99.3} \\
\ours (\emph{ours}) & \textbf{90.7} & \textbf{99.5} & \underline{70.5} & \textbf{97.4} & \textbf{97.5} & \textbf{100.0} & \textbf{91.2} & \textbf{100.0} & \textbf{74.2} & \textbf{100.0} & \textbf{82.0} & \textbf{99.3} \\
\bottomrule
\end{tabular}
\end{table}

\para{Quantitative results on KITTI and nuScenes.}
We first compare against $3$ baselines, \ie, GeoTrans~\cite{qin2022geometric}, BUFFER~\cite{ao2023buffer}, and PARENet~\cite{yao2024pare}, all trained on KITTI.
As shown in \cref{table:results-outdoor}~(top), \ours outperforms all baselines on both benchmarks, \eg, by $6.6$ pp in \emph{Inlier Ratio} on KITTI, indicating that cross-domain training yields more discriminative features and generates more accurate correspondences.
On nuScenes, despite a lower \emph{Inlier Ratio}, our method achieves substantially higher \emph{Registration Recall}, demonstrating stronger robustness to difficult cases.
Additionally, we report results for \ours trained without nuScenes (marked by $^{\diamond}$). This variant achieves higher \emph{Registration Recall} than the baselines on most benchmarks, except MIT, further validating our method. Because both nuScenes and MIT employ LiDAR sensors with fewer beams than KITTI, omitting nuScenes during training slightly degrades performance relative to the model trained with nuScenes.

\para{Quantitative results on unseen benchmarks.}
We further assess the generalizability on $4$ unseen outdoor benchmarks, \ie, ETH, WOD, MIT and Oxford. As these benchmarks all adopts spinning LiDAR scanners, most baselines except PARENet generalize reasonably well.
\ours achieves the best performance on all metrics across all benchmarks, which again underscoring the advantage of cross-domain training for robust generalizability.
Note that the outdoor-only \ours (marked by $^{\star}$) trained on KITTI and nuScenes outperforms the hybrid model without nuScenes on most benchmarks, suggesting that domain-relevant training data remain important for performance within each domain.

\subsection{Experiments on Object Benchmarks}
\label{sec:exp:object}


\begin{table}[!t]
\scriptsize
\centering
\caption{
Results on object benchmarks.
\textbf{Boldfaced} numbers highlight the best and the second best are \underline{underlined}.
$^{\dagger}$ and $^{\ddagger}$ denote the models pretrained on 3DMatch and KITTI, respectively.
$^{\star}$ denotes the model trained on object data.
}
\label{table:results-object}
\begin{tabular}{lcccccccc}
\toprule
\multirow{2}{*}{Model} & \multicolumn{4}{c}{ModelNet} & \multicolumn{4}{c}{ShapeNet} \\
\cmidrule(r){2-9}
 & IR(\%) & RR(\%) & RRE($^{\circ}$) & RTE(m) & IR(\%) & RR(\%) & RRE($^{\circ}$) & RTE(m) \\
\midrule
RPM-Net~\cite{yew2020rpm} & - & 90.9 & 2.412 & 0.030 & - & 70.2 & 6.177 & 0.058 \\
Predator~\cite{huang2021predator} & 62.8 & 90.1 & 2.614 & 0.033 & 56.0 & 73.6 & 5.668 & 0.067 \\
CoFiNet~\cite{yu2021cofinet} & 48.2 & \underline{91.8} & 2.802 & 0.038 & 45.7 & 73.5 & 6.834 & 0.068 \\
GeoTrans~\cite{qin2022geometric} & \textbf{73.3} & 90.4 & 2.636 & 0.032 & \textbf{67.3} & \underline{76.0} & \textbf{4.492} & \underline{0.053} \\
\midrule
BUFFER-X$^{\dagger}$~\cite{seo2025buffer} & 25.3 & 13.3 & 95.483 & 0.608 & 50.9 & 39.3 & 67.315 & 0.396 \\
BUFFER-X$^{\ddagger}$~\cite{seo2025buffer} & 46.4 & 34.9 & 62.221 & 0.401 & 65.0 & 55.3 & 47.209 & 0.276 \\
\midrule
\ours{}$^{\star}$ (\emph{ours}) & \underline{66.4} & 90.9 & \underline{2.026} & \underline{0.026} & 65.6 & \underline{76.0} & 4.684 & \underline{0.053} \\
\ours (\emph{ours}) & 65.6 & \textbf{94.2} & \textbf{1.875} & \textbf{0.023} & \underline{65.9} & \textbf{79.6} & \underline{4.592} & \textbf{0.047} \\
\bottomrule
\end{tabular}
\end{table}

We compare with $5$ baselines: RPM-Net~\cite{yew2020rpm}, Predator~\cite{huang2021predator}, CoFiNet~\cite{yu2021cofinet}, GeoTrans~\cite{qin2022geometric}, and BUFFER-X~\cite{ao2023buffer}, where the first $4$ baselines are trained on ModelNet, while BUFFER-X uses the weights pretrained on 3DMatch and KITTI.
As shown in \cref{table:results-object}, our method consistently outperforms all object baselines by a large margin on both ModelNet and ShapeNet.
We notice that GeoTrans attains better results in \emph{Inlier Ratio}. This is because object point clouds are usually small, so the global geometric structure embedding is less affected by the voxelization noises and contributes to better feature representation in most cases. However, thanks to the rich geometric patterns from diverse domains, \ours extracts more accurate correspondences in the harder cases, thus contributing to better \emph{Registration Recall}.
Another important observation is that both variants of BUFFER-X fail to achieve reasonable results on the object registration benchmarks, which unveils the significant difficulty of generalization from scene registration to object registration due to the severe geometric discrepancy. In contrast, benefiting from the diverse data from various domains during training, our method achieves strong generalizability across a wide range of domains.

\subsection{Ablation Studies}
\label{sec:exp:ablation}

At last, we conduct extensive ablation studies to better understand the design choices of \ours on the $7$ unseen benchmarks, \ie, ScanNet++, TIERS, ETH, WOD, MIT, Oxford and ShapeNet.


\begin{table}[!t]
\setlength{\tabcolsep}{3pt}
\scriptsize
\centering
\caption{
Ablation studies.
\textbf{Boldfaced} numbers highlight the best and the second best are \underline{underlined}.
\textbf{OS}: overlap scores.
\textbf{O\&T}: overlap scores and transformer.
\textbf{GN}: geometry-aware normalization.
}
\label{table:results-ablation}
\begin{tabular}{lcccccccccccccc}
\toprule
\multirow{2}{*}{Model} & \multicolumn{2}{c}{ScanNet++} & \multicolumn{2}{c}{TIERS} & \multicolumn{2}{c}{ETH} & \multicolumn{2}{c}{WOD} & \multicolumn{2}{c}{MIT} & \multicolumn{2}{c}{Oxford} & \multicolumn{2}{c}{ShapeNet} \\
\cmidrule(r){2-15}
 & IR(\%) & RR(\%) & IR(\%) & RR(\%) & IR(\%) & RR(\%) & IR(\%) & RR(\%) & IR(\%) & RR(\%) & IR(\%) & RR(\%) & IR(\%) & RR(\%) \\
\midrule
(a.1) hybrid$^{\star}$ & \textbf{70.6} & \textbf{90.0} & \textbf{66.0} & \textbf{90.0} & \textbf{97.5} & \textbf{100.0} & \textbf{91.2} & \textbf{100.0} & \textbf{74.2} & \textbf{100.0} & \textbf{82.0} & \textbf{99.3} & \textbf{65.9} & \textbf{79.6} \\
(a.2) indoor & \underline{65.9} & \underline{89.2} & 48.5 & 81.3 & 77.9 & 87.3 & 66.2 & 98.5 & 23.8 & 58.3 & 64.4 & 96.7 & 21.0 & 30.7 \\
(a.3) outdoor & 41.1 & 72.5 & \underline{62.3} & \underline{87.5} & \underline{93.8} & \underline{97.9} & \underline{90.4} & \textbf{100.0} & \underline{72.3} & \underline{98.7} & \underline{79.2} & \underline{99.0} & 21.0 & 28.8 \\
(a.4) object & 28.9 & 56.0 & 42.7 & 68.4 & 32.3 & 39.5 & 36.2 & 66.2 & 31.3 & 75.2 & 52.0 & 86.7 & \underline{65.6} & \underline{76.0} \\
\midrule
(b.1) w/ OS$^{\star}$ & \textbf{70.6} & \textbf{90.0} & \textbf{66.0} & \textbf{90.0} & \textbf{97.5} & \textbf{100.0} & \textbf{91.2} & \textbf{100.0} & \textbf{74.2} & \textbf{100.0} & \textbf{82.0} & \textbf{99.3} & \textbf{65.9} & 79.6 \\
(b.2) w/o OS & 68.3 & 88.9 & 64.7 & 89.1 & 96.5 & 99.4 & 90.8 & \textbf{100.0} & 72.9 & 99.1 & 80.9 & \textbf{99.3} & 63.6 & \textbf{80.3} \\
(b.3) w/o O\&T & 63.7 & 88.9 & 61.0 & 88.7 & 96.3 & 99.0 & 86.0 & \textbf{100.0} & 69.3 & 99.1 & 74.9 & 99.0 & 58.6 & 79.6 \\
\midrule
(c.1) GN$^{\star}$ & \textbf{70.6} & 90.0 & \textbf{66.0} & \textbf{90.0} & \textbf{97.5} & \textbf{100.0} & 91.2 & \textbf{100.0} & \textbf{74.2} & \textbf{100.0} & 82.0 & \textbf{99.3} & \textbf{65.9} & \textbf{79.6} \\
(c.2) fixed & 66.8 & \textbf{90.6} & 47.8 & 89.2 & 62.4 & 99.0 & \textbf{91.3} & \textbf{100.0} & 72.0 & 99.1 & \textbf{86.0} & \textbf{99.3} & 65.9 & 79.6 \\
\bottomrule
\end{tabular}
\end{table}

\para{Cross-domain training.}
We first investigate how cross-domain training affects the generalizability, as shown in \cref{table:results-ablation}~(a). We compare $4$ variants of \ours, respectively trained on (1) hybrid data, (2) indoor data, (3) outdoor data, and (4) object data.
For \emph{Registration Recall}, the indoor model performs quite well on most outdoor benchmarks except MIT. The outdoor model performs well on TIERS but fails to generalize to ScanNet++.
Both models fail on the object benchmark, while the object model also fails on the scene-level benchmarks.
For \emph{Inlier Ratio}, all models suffer from significant performance drop on different domains.
The hybrid model achieves the best performance on all benchmarks, demonstrating the efficacy of cross-domain training.

\para{Point matching.}
We then study the efficacy of the designs in the point matching stage in \cref{table:results-ablation}~(b). We compare $3$ variants of \ours: (1) the full model, (2) the model without the overlap scores, and (3) the model further without the geometric transformer.
By predicting the overlapping scores, the model suppresses the non-overlapping points during matching and thus yields more accurate correspondences. And removing the transformer causes a significant performance degradation, further confirming the importance of the inter-point-cloud information exchange in point matching.

\para{Geometry-aware normalization.}
We evaluate the efficacy of our geometry-aware normalization by comparing with the model with static voxel sizes in \cref{table:results-ablation}~(c). Following~\cite{ao2023buffer,qin2022geometric}, we set the voxel size to $2.5$cm for indoor and object benchmarks, and $30$cm for outdoor benchmarks. Both models use the same weights trained on static voxel sizes. The model with geometry-aware normalization consistently outperforms the static counterpart on most benchmarks.
Note that the benefit of our design is not limited to the average improvement. It eliminates the need to manually select a domain-specific voxel size at inference time. This is particularly important in practical applications where an agent may frequently transition between environments with substantially different geometric scales.

\subsection{Qualitative Results}

\begin{figure*}[t]
  \centering
  \begin{overpic}[width=\linewidth]{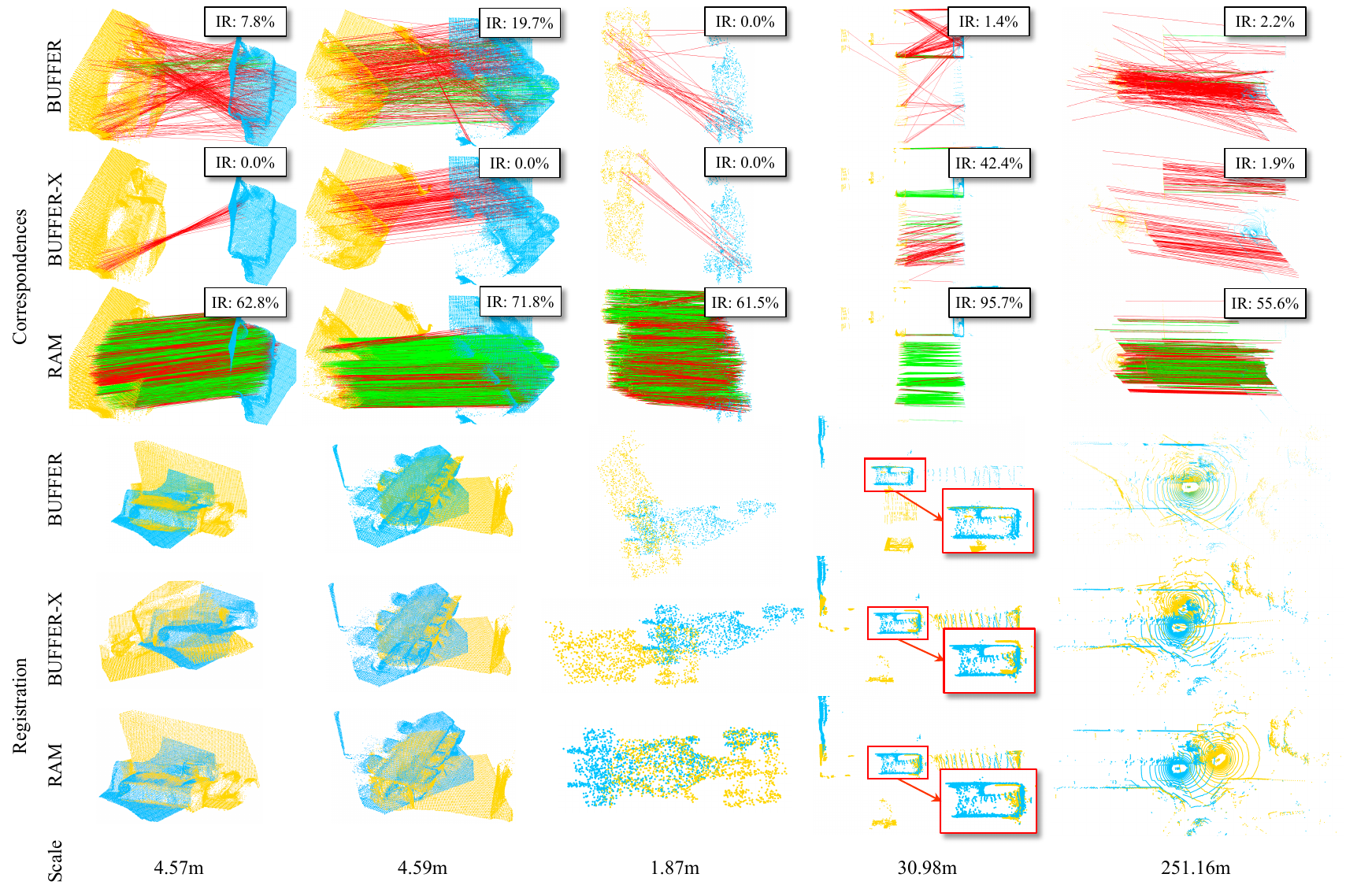}
  \end{overpic}
  \caption{Visualization of the registration results on various benchmarks.}
  \label{fig:gallery}
\end{figure*}

\cref{fig:gallery} visualizes some registration results of BUFFER, BUFFER-X and \ours on various benchmarks. Our method can effectively extract accurate correspondences across different domains, despite significant geometric discrepancy in terms of scene scales, point densities, structural patterns. And our method is still robust in case of incomplete geometry, duplicated objects, and symmetric structures.

\section{Conclusion}
\label{sec:conclusion}

In this paper, we investigated cross-domain training for point cloud registration and introduced a unified registration model, \ours. Extensive experiments across a broad set of benchmarks demonstrate that \ours delivers strong cross-domain generalizability and robustness, achieving state-of-the-art performance in diverse indoor, outdoor, and object registration scenarios.

Despite these advances, \ours can be improved in several respects.
First, the KPConv-based backbone is still rotation-variant. Although \ours already shows promising robustness to large rotations as demonstrated in the appendix, it may still struggle under extreme scenarios. We assessed recent rotation-equivariant backbones, but their memory footprint currently prevents processing large-scale point clouds within a unified architecture.
Second, \ours may not scale to extremely large point clouds, \eg, urban-scale UAV captures. A practical remedy is to partition scenes into manageable tiles and process them separately, potentially with overlap and streaming.
Third, we observe that distant outliers in point clouds can inflate the initial voxel size and oversmooth geometry. This could be mitigated by integrating a separate point cloud outlier removal step.

Beyond registration, \ours can serve as a general-purpose model for point cloud processing, and we hope it will help catalyze future research toward point cloud foundation models.

\section{Acknowledgements}
\label{sec:ack}

This work is supported in part by the NSFC (62402516, 62325211, 62132021, 62402505) and the Research Fund of Jiangsu Key Laboratory of AI for Industries (E6420016G8).

{\small
\bibliographystyle{unsrtnat}
\bibliography{main}

\begin{thebibliography}{54}
\providecommand{\natexlab}[1]{#1}
\providecommand{\url}[1]{\texttt{#1}}
\expandafter\ifx\csname urlstyle\endcsname\relax
  \providecommand{\doi}[1]{doi: #1}\else
  \providecommand{\doi}{doi: \begingroup \urlstyle{rm}\Url}\fi

\bibitem[Choy et~al.(2019{\natexlab{a}})Choy, Park, and Koltun]{choy2019fully}
Christopher Choy, Jaesik Park, and Vladlen Koltun.
\newblock Fully convolutional geometric features.
\newblock In \emph{CVPR}, pages 8958--8966, 2019{\natexlab{a}}.

\bibitem[Huang et~al.(2021{\natexlab{a}})Huang, Gojcic, Usvyatsov, Wieser, and
  Schindler]{huang2021predator}
Shengyu Huang, Zan Gojcic, Mikhail Usvyatsov, Andreas Wieser, and Konrad
  Schindler.
\newblock Predator: Registration of 3d point clouds with low overlap.
\newblock In \emph{CVPR}, pages 4267--4276, 2021{\natexlab{a}}.

\bibitem[Qin et~al.(2022)Qin, Yu, Wang, Guo, Peng, and Xu]{qin2022geometric}
Zheng Qin, Hao Yu, Changjian Wang, Yulan Guo, Yuxing Peng, and kai Xu.
\newblock Geometric transformer for fast and robust point cloud registration.
\newblock In \emph{CVPR}, pages 11143--11152, 2022.

\bibitem[Yu et~al.(2023{\natexlab{a}})Yu, Qin, Hou, Saleh, Li, Busam, and
  Ilic]{yu2023rotation}
Hao Yu, Zheng Qin, Ji~Hou, Mahdi Saleh, Dongsheng Li, Benjamin Busam, and
  Slobodan Ilic.
\newblock Rotation-invariant transformer for point cloud matching.
\newblock In \emph{CVPR}, pages 5384--5393, 2023{\natexlab{a}}.

\bibitem[Ao et~al.(2023)Ao, Hu, Wang, Xu, and Guo]{ao2023buffer}
Sheng Ao, Qingyong Hu, Hanyun Wang, Kai Xu, and Yulan Guo.
\newblock Buffer: Balancing accuracy, efficiency, and generalizability in point
  cloud registration.
\newblock In \emph{CVPR}, pages 1255--1264, 2023.

\bibitem[Yew and Lee(2020)]{yew2020rpm}
Zi~Jian Yew and Gim~Hee Lee.
\newblock Rpm-net: Robust point matching using learned features.
\newblock In \emph{CVPR}, pages 11824--11833, 2020.

\bibitem[Zeng et~al.(2017)Zeng, Song, Nie{\ss}ner, Fisher, Xiao, and
  Funkhouser]{zeng20173dmatch}
Andy Zeng, Shuran Song, Matthias Nie{\ss}ner, Matthew Fisher, Jianxiong Xiao,
  and Thomas Funkhouser.
\newblock 3dmatch: Learning local geometric descriptors from rgb-d
  reconstructions.
\newblock In \emph{CVPR}, pages 1802--1811, 2017.

\bibitem[Dai et~al.(2017)Dai, Chang, Savva, Halber, Funkhouser, and
  Nie{\ss}ner]{dai2017scannet}
Angela Dai, Angel~X Chang, Manolis Savva, Maciej Halber, Thomas Funkhouser, and
  Matthias Nie{\ss}ner.
\newblock Scannet: Richly-annotated 3d reconstructions of indoor scenes.
\newblock In \emph{CVPR}, pages 5828--5839, 2017.

\bibitem[Geiger et~al.(2012)Geiger, Lenz, and Urtasun]{geiger2012we}
Andreas Geiger, Philip Lenz, and Raquel Urtasun.
\newblock Are we ready for autonomous driving? the kitti vision benchmark
  suite.
\newblock In \emph{CVPR}, pages 3354--3361. IEEE, 2012.

\bibitem[Caesar et~al.(2020)Caesar, Bankiti, Lang, Vora, Liong, Xu, Krishnan,
  Pan, Baldan, and Beijbom]{caesar2020nuscenes}
Holger Caesar, Varun Bankiti, Alex~H Lang, Sourabh Vora, Venice~Erin Liong,
  Qiang Xu, Anush Krishnan, Yu~Pan, Giancarlo Baldan, and Oscar Beijbom.
\newblock nuscenes: A multimodal dataset for autonomous driving.
\newblock In \emph{CVPR}, pages 11621--11631, 2020.

\bibitem[Wu et~al.(2015)Wu, Song, Khosla, Yu, Zhang, Tang, and Xiao]{wu20153d}
Zhirong Wu, Shuran Song, Aditya Khosla, Fisher Yu, Linguang Zhang, Xiaoou Tang,
  and Jianxiong Xiao.
\newblock 3d shapenets: A deep representation for volumetric shapes.
\newblock In \emph{CVPR}, pages 1912--1920, 2015.

\bibitem[Chang et~al.(2015)Chang, Funkhouser, Guibas, Hanrahan, Huang, Li,
  Savarese, Savva, Song, Su, et~al.]{chang2015shapenet}
Angel~X Chang, Thomas Funkhouser, Leonidas Guibas, Pat Hanrahan, Qixing Huang,
  Zimo Li, Silvio Savarese, Manolis Savva, Shuran Song, Hao Su, et~al.
\newblock Shapenet: An information-rich 3d model repository.
\newblock \emph{arXiv preprint arXiv:1512.03012}, 2015.

\bibitem[Seo et~al.(2025)Seo, Lim, Lee, Carlone, and Park]{seo2025buffer}
Minkyun Seo, Hyungtae Lim, Kanghee Lee, Luca Carlone, and Jaesik Park.
\newblock Buffer-x: Towards zero-shot point cloud registration in diverse
  scenes.
\newblock In \emph{ICCV}, pages 3851--3862, 2025.

\bibitem[Kirillov et~al.(2023)Kirillov, Mintun, Ravi, Mao, Rolland, Gustafson,
  Xiao, Whitehead, Berg, Lo, et~al.]{kirillov2023segment}
Alexander Kirillov, Eric Mintun, Nikhila Ravi, Hanzi Mao, Chloe Rolland, Laura
  Gustafson, Tete Xiao, Spencer Whitehead, Alexander~C Berg, Wan-Yen Lo, et~al.
\newblock Segment anything.
\newblock In \emph{ICCV}, pages 4015--4026, 2023.

\bibitem[Oquab et~al.(2023)Oquab, Darcet, Moutakanni, Vo, Szafraniec, Khalidov,
  Fernandez, Haziza, Massa, El-Nouby, et~al.]{oquab2023dinov2}
Maxime Oquab, Timoth{\'e}e Darcet, Th{\'e}o Moutakanni, Huy Vo, Marc
  Szafraniec, Vasil Khalidov, Pierre Fernandez, Daniel Haziza, Francisco Massa,
  Alaaeldin El-Nouby, et~al.
\newblock Dinov2: Learning robust visual features without supervision.
\newblock \emph{arXiv preprint arXiv:2304.07193}, 2023.

\bibitem[Yang et~al.(2024)Yang, Kang, Huang, Zhao, Xu, Feng, and
  Zhao]{yang2024depth}
Lihe Yang, Bingyi Kang, Zilong Huang, Zhen Zhao, Xiaogang Xu, Jiashi Feng, and
  Hengshuang Zhao.
\newblock Depth anything v2.
\newblock \emph{NeurIPS}, 37:\penalty0 21875--21911, 2024.

\bibitem[Thomas et~al.(2019)Thomas, Qi, Deschaud, Marcotegui, Goulette, and
  Guibas]{thomas2019kpconv}
Hugues Thomas, Charles~R Qi, Jean-Emmanuel Deschaud, Beatriz Marcotegui,
  Fran{\c{c}}ois Goulette, and Leonidas~J Guibas.
\newblock Kpconv: Flexible and deformable convolution for point clouds.
\newblock In \emph{ICCV}, pages 6411--6420, 2019.

\bibitem[Choy et~al.(2019{\natexlab{b}})Choy, Gwak, and Savarese]{choy20194d}
Christopher Choy, JunYoung Gwak, and Silvio Savarese.
\newblock 4d spatio-temporal convnets: Minkowski convolutional neural networks.
\newblock In \emph{CVPR}, pages 3075--3084, 2019{\natexlab{b}}.

\bibitem[Ao et~al.(2021)Ao, Hu, Yang, Markham, and Guo]{ao2021spinnet}
Sheng Ao, Qingyong Hu, Bo~Yang, Andrew Markham, and Yulan Guo.
\newblock Spinnet: Learning a general surface descriptor for 3d point cloud
  registration.
\newblock In \emph{CVPR}, pages 11753--11762, 2021.

\bibitem[Yao et~al.(2024)Yao, Du, Cui, Tang, and Yang]{yao2024pare}
Runzhao Yao, Shaoyi Du, Wenting Cui, Canhui Tang, and Chengwu Yang.
\newblock Pare-net: Position-aware rotation-equivariant networks for robust
  point cloud registration.
\newblock In \emph{ECCV}, pages 287--303. Springer, 2024.

\bibitem[Yu et~al.(2021)Yu, Li, Saleh, Busam, and Ilic]{yu2021cofinet}
Hao Yu, Fu~Li, Mahdi Saleh, Benjamin Busam, and Slobodan Ilic.
\newblock Cofinet: Reliable coarse-to-fine correspondences for robust
  pointcloud registration.
\newblock \emph{NeurIPS}, 34:\penalty0 23872--23884, 2021.

\bibitem[Wang et~al.(2022)Wang, Liu, Dong, and Wang]{wang2022you}
Haiping Wang, Yuan Liu, Zhen Dong, and Wenping Wang.
\newblock You only hypothesize once: Point cloud registration with
  rotation-equivariant descriptors.
\newblock In \emph{ACM MM}, pages 1630--1641, 2022.

\bibitem[Huang et~al.(2022)Huang, Qu, Zuo, Fang, and Zhao]{huang2022imfnet}
Xiaoshui Huang, Wentao Qu, Yifan Zuo, Yuming Fang, and Xiaowei Zhao.
\newblock Imfnet: Interpretable multimodal fusion for point cloud registration.
\newblock \emph{RAL}, 7\penalty0 (4):\penalty0 12323--12330, 2022.

\bibitem[Huang et~al.(2025)Huang, Huang, Zuo, Gong, Zhang, Liu, and
  Fang]{huang2025psreg}
Xiaoshui Huang, Zhou Huang, Yifan Zuo, Yongshun Gong, Chengdong Zhang, Deyang
  Liu, and Yuming Fang.
\newblock Psreg: Prior-guided sparse mixture of experts for point cloud
  registration.
\newblock In \emph{AAAI}, volume~39, pages 3788--3796, 2025.

\bibitem[Bai et~al.(2020)Bai, Luo, Zhou, Fu, Quan, and Tai]{bai2020d3feat}
Xuyang Bai, Zixin Luo, Lei Zhou, Hongbo Fu, Long Quan, and Chiew-Lan Tai.
\newblock D3feat: Joint learning of dense detection and description of 3d local
  features.
\newblock In \emph{CVPR}, pages 6359--6367, 2020.

\bibitem[Fischler and Bolles(1981)]{fischler1981random}
Martin~A Fischler and Robert~C Bolles.
\newblock Random sample consensus: a paradigm for model fitting with
  applications to image analysis and automated cartography.
\newblock \emph{Communications of the ACM}, 24\penalty0 (6):\penalty0 381--395,
  1981.

\bibitem[Chen et~al.(2022)Chen, Sun, Yang, and Tao]{chen2022sc2}
Zhi Chen, Kun Sun, Fan Yang, and Wenbing Tao.
\newblock Sc2-pcr: A second order spatial compatibility for efficient and
  robust point cloud registration.
\newblock In \emph{CVPR}, pages 13221--13231, 2022.

\bibitem[Zhang et~al.(2023)Zhang, Yang, Zhang, and Zhang]{zhang20233d}
Xiyu Zhang, Jiaqi Yang, Shikun Zhang, and Yanning Zhang.
\newblock 3d registration with maximal cliques.
\newblock In \emph{CVPR}, pages 17745--17754, 2023.

\bibitem[Choy et~al.(2020)Choy, Dong, and Koltun]{choy2020deep}
Christopher Choy, Wei Dong, and Vladlen Koltun.
\newblock Deep global registration.
\newblock In \emph{CVPR}, pages 2514--2523, 2020.

\bibitem[Bai et~al.(2021)Bai, Luo, Zhou, Chen, Li, Hu, Fu, and
  Tai]{bai2021pointdsc}
Xuyang Bai, Zixin Luo, Lei Zhou, Hongkai Chen, Lei Li, Zeyu Hu, Hongbo Fu, and
  Chiew-Lan Tai.
\newblock Pointdsc: Robust point cloud registration using deep spatial
  consistency.
\newblock In \emph{CVPR}, pages 15859--15869, 2021.

\bibitem[Aoki et~al.(2019)Aoki, Goforth, Srivatsan, and
  Lucey]{aoki2019pointnetlk}
Yasuhiro Aoki, Hunter Goforth, Rangaprasad~Arun Srivatsan, and Simon Lucey.
\newblock Pointnetlk: Robust \& efficient point cloud registration using
  pointnet.
\newblock In \emph{CVPR}, pages 7163--7172, 2019.

\bibitem[Fu et~al.(2021)Fu, Liu, Luo, and Wang]{fu2021robust}
Kexue Fu, Shaolei Liu, Xiaoyuan Luo, and Manning Wang.
\newblock Robust point cloud registration framework based on deep graph
  matching.
\newblock In \emph{CVPR}, pages 8893--8902, 2021.

\bibitem[Xu et~al.(2021{\natexlab{a}})Xu, Liu, Wang, Liu, and
  Zeng]{xu2021omnet}
Hao Xu, Shuaicheng Liu, Guangfu Wang, Guanghui Liu, and Bing Zeng.
\newblock Omnet: Learning overlapping mask for partial-to-partial point cloud
  registration.
\newblock In \emph{ICCV}, pages 3132--3141, 2021{\natexlab{a}}.

\bibitem[Huang et~al.(2017{\natexlab{a}})Huang, Zhang, Fan, Wu, and
  Yuan]{huang2017systematic}
Xiaoshui Huang, Jian Zhang, Lixin Fan, Qiang Wu, and Chun Yuan.
\newblock A systematic approach for cross-source point cloud registration by
  preserving macro and micro structures.
\newblock \emph{IEEE TIP}, 26\penalty0 (7):\penalty0 3261--3276,
  2017{\natexlab{a}}.

\bibitem[Huang et~al.(2017{\natexlab{b}})Huang, Zhang, Wu, Fan, and
  Yuan]{huang2017coarse}
Xiaoshui Huang, Jian Zhang, Qiang Wu, Lixin Fan, and Chun Yuan.
\newblock A coarse-to-fine algorithm for matching and registration in 3d
  cross-source point clouds.
\newblock \emph{IEEE TCSVT}, 28\penalty0 (10):\penalty0 2965--2977,
  2017{\natexlab{b}}.

\bibitem[Huang et~al.(2023)Huang, Mei, and Zhang]{huang2023cross}
Xiaoshui Huang, Guofeng Mei, and Jian Zhang.
\newblock Cross-source point cloud registration: Challenges, progress and
  prospects.
\newblock \emph{Neurocomputing}, 548:\penalty0 126383, 2023.

\bibitem[Huang et~al.(2021{\natexlab{b}})Huang, Mei, Zhang, and
  Abbas]{huang2103comprehensive}
X~Huang, G~Mei, J~Zhang, and R~Abbas.
\newblock A comprehensive survey on point cloud registration. arxiv 2021.
\newblock \emph{arXiv preprint arXiv:2103.02690}, 2021{\natexlab{b}}.

\bibitem[Zhao et~al.(2024)Zhao, Du, Guo, and Ma]{zhao2024vrhcf}
Guiyu Zhao, Zewen Du, Zhentao Guo, and Hongbin Ma.
\newblock Vrhcf: Cross-source point cloud registration via voxel representation
  and hierarchical correspondence filtering.
\newblock In \emph{ICME}, pages 1--6. IEEE, 2024.

\bibitem[Zhao et~al.(2025)Zhao, Guo, Du, and Ma]{zhao2025cross}
Guiyu Zhao, Zhentao Guo, Zewen Du, and Hongbin Ma.
\newblock Cross-pcr: A robust cross-source point cloud registration framework.
\newblock In \emph{AAAI}, volume~39, pages 10403--10411, 2025.

\bibitem[Xu et~al.(2021{\natexlab{b}})Xu, Ding, Zhao, and Qi]{xu2021paconv}
Mutian Xu, Runyu Ding, Hengshuang Zhao, and Xiaojuan Qi.
\newblock Paconv: Position adaptive convolution with dynamic kernel assembling
  on point clouds.
\newblock In \emph{CVPR}, pages 3173--3182, 2021{\natexlab{b}}.

\bibitem[Alexiou et~al.(2021)Alexiou, Zhou, Viola, and
  Cesar]{alexiou2021pointpca}
Evangelos Alexiou, Xuemei Zhou, Irene Viola, and Pablo Cesar.
\newblock Pointpca: Point cloud objective quality assessment using pca-based
  descriptors.
\newblock \emph{arXiv preprint arXiv:2111.12663}, 2021.

\bibitem[Yu et~al.(2023{\natexlab{b}})Yu, Ren, Zhang, Zhou, Lin, and
  Dai]{yu2023peal}
Junle Yu, Luwei Ren, Yu~Zhang, Wenhui Zhou, Lili Lin, and Guojun Dai.
\newblock Peal: Prior-embedded explicit attention learning for low-overlap
  point cloud registration.
\newblock In \emph{CVPR}, pages 17702--17711, 2023{\natexlab{b}}.

\bibitem[Shah et~al.(2024)Shah, Bikshandi, Zhang, Thakkar, Ramani, and
  Dao]{shah2024flashattention}
Jay Shah, Ganesh Bikshandi, Ying Zhang, Vijay Thakkar, Pradeep Ramani, and Tri
  Dao.
\newblock Flashattention-3: Fast and accurate attention with asynchrony and
  low-precision.
\newblock \emph{NeurIPS}, 37:\penalty0 68658--68685, 2024.

\bibitem[Wang et~al.(2020)Wang, Li, Khabsa, Fang, and Ma]{wang2020linformer}
Sinong Wang, Belinda~Z Li, Madian Khabsa, Han Fang, and Hao Ma.
\newblock Linformer: Self-attention with linear complexity.
\newblock \emph{arXiv preprint arXiv:2006.04768}, 2020.

\bibitem[Yang et~al.(2022)Yang, Guo, Chen, and Tao]{yang2022one}
Fan Yang, Lin Guo, Zhi Chen, and Wenbing Tao.
\newblock One-inlier is first: Towards efficient position encoding for point
  cloud registration.
\newblock \emph{NeurIPS}, 35:\penalty0 6982--6995, 2022.

\bibitem[Lindenberger et~al.(2023)Lindenberger, Sarlin, and
  Pollefeys]{lindenberger2023lightglue}
Philipp Lindenberger, Paul-Edouard Sarlin, and Marc Pollefeys.
\newblock Lightglue: Local feature matching at light speed.
\newblock In \emph{ICCV}, pages 17627--17638, 2023.

\bibitem[Yeshwanth et~al.(2023)Yeshwanth, Liu, Nie{\ss}ner, and
  Dai]{yeshwanth2023scannet++}
Chandan Yeshwanth, Yueh-Cheng Liu, Matthias Nie{\ss}ner, and Angela Dai.
\newblock Scannet++: A high-fidelity dataset of 3d indoor scenes.
\newblock In \emph{ICCV}, pages 12--22, 2023.

\bibitem[Li et~al.(2022)Li, Yu, Pena~Queralta, and Westerlund]{li2022multi}
Qingqing Li, Xianjia Yu, Jorge Pena~Queralta, and Tomi Westerlund.
\newblock Multi-modal lidar dataset for benchmarking general-purpose
  localization and mapping algorithms.
\newblock In \emph{IROS}, pages 3837--3844. IEEE, 2022.

\bibitem[Pomerleau et~al.(2012)Pomerleau, Liu, Colas, and
  Siegwart]{pomerleau2012challenging}
Fran{\c{c}}ois Pomerleau, Ming Liu, Francis Colas, and Roland Siegwart.
\newblock Challenging data sets for point cloud registration algorithms.
\newblock \emph{The International Journal of Robotics Research}, 31\penalty0
  (14):\penalty0 1705--1711, 2012.

\bibitem[Tian et~al.(2023)Tian, Chang, Quang, Schang, Nieto-Granda, How, and
  Carlone]{tian2023resilient}
Yulun Tian, Yun Chang, Long Quang, Arthur Schang, Carlos Nieto-Granda,
  Jonathan~P How, and Luca Carlone.
\newblock Resilient and distributed multi-robot visual slam: Datasets,
  experiments, and lessons learned.
\newblock In \emph{IROS}, pages 11027--11034. IEEE, 2023.

\bibitem[Sun et~al.(2020)Sun, Kretzschmar, Dotiwalla, Chouard, Patnaik, Tsui,
  Guo, Zhou, Chai, Caine, et~al.]{sun2020scalability}
Pei Sun, Henrik Kretzschmar, Xerxes Dotiwalla, Aurelien Chouard, Vijaysai
  Patnaik, Paul Tsui, James Guo, Yin Zhou, Yuning Chai, Benjamin Caine, et~al.
\newblock Scalability in perception for autonomous driving: Waymo open dataset.
\newblock In \emph{CVPR}, pages 2446--2454, 2020.

\bibitem[Liu et~al.(2024)Liu, Zhu, Wang, Zhou, Chang, and Guo]{liu2024extend}
Quan Liu, Hongzi Zhu, Zhenxi Wang, Yunsong Zhou, Shan Chang, and Minyi Guo.
\newblock Extend your own correspondences: Unsupervised distant point cloud
  registration by progressive distance extension.
\newblock In \emph{CVPR}, pages 20816--20826, 2024.

\bibitem[Ramezani et~al.(2020)Ramezani, Wang, Camurri, Wisth, Mattamala, and
  Fallon]{ramezani2020newer}
Milad Ramezani, Yiduo Wang, Marco Camurri, David Wisth, Matias Mattamala, and
  Maurice Fallon.
\newblock The newer college dataset: Handheld lidar, inertial and vision with
  ground truth.
\newblock In \emph{IROS}, pages 4353--4360. IEEE, 2020.

\bibitem[Wang et~al.(2023)Wang, Liu, Hu, Wang, Chen, Dong, Guo, Wang, and
  Yang]{wang2023roreg}
Haiping Wang, Yuan Liu, Qingyong Hu, Bing Wang, Jianguo Chen, Zhen Dong, Yulan
  Guo, Wenping Wang, and Bisheng Yang.
\newblock Roreg: Pairwise point cloud registration with oriented descriptors
  and local rotations.
\newblock \emph{IEEE TPAMI}, 45\penalty0 (8):\penalty0 10376--10393, 2023.

\end{thebibliography}
}




\appendix
\section{Appendix}
\label{sec:appendix}

\subsection{Implementation Details}

\para{Voxel size initialization.}
We set the reference voxel size $v_{\text{ref}} = 0.025\text{m}$, a conventional choice for 3DMatch. And we define $\phi(\cdot)$ as a monotonically increasing piecewise function:
\begin{equation}
\phi(x) = \begin{cases}
0.1 \log_2\left(\dfrac{x}{25}\right), & x \geq 50 \\[1.2em]
0.001\,(x - 25) + 0.075, & 25 \leq x < 50 \\[1.2em]
0.0025\,(x - 1) + 0.015, & x < 25
\end{cases}.
\end{equation} 
This function is designed based on several empirical anchor points, \ie, $\phi(1) = 0.015\text{m}$, $\phi(5) = 0.025\text{m}$, $\phi(50) = 0.1\text{m}$ and $\phi(200) = 0.3\text{m}$. For point clouds smaller than $25$m, the environments range from rooms to halls or buildings, so we want the initial voxel size to increase rapidly to reflect this change. 
For point clouds larger than $50$m, which typically correspond to outdoor environments, an excessively large voxel size can impair registration accuracy; therefore, we employ a logarithmic function to prevent the voxel size from growing too quickly.
We sample $512$ seeds for each point cloud, and collect the neighbors within $8 v^{(t)}$ for each seed to compute the geometric properties.
Moreover, we set the relative error threshold $\tau = 0.15$ and the scaling factor $\alpha = 0.05$. The normalization process is iterated for up to $T=5$ times.

\para{Model details.}
We use a $4$-stage KPConv-FPN backbone, with the initial voxel size of $0.025$m. The point cloud resolution is halved after each stage, while the feature channels are doubled. The search radius of KPConv is $2.5{\times}$ of the voxel size of each stage. For \egtr, we use $128$ neighbors for each superpoint to compute the geometric structure embedding. For other hyperparameters, we follow the 3DMatch settings in GeoTrans~\cite{qin2022geometric}.

\para{Training and testing.}
Our model is implemented and trained with an Intel i7-12700K CPU and an NVIDIA RTX 5090 GPU. We use AdamW optimizer to train all models for $3$M iterations, with the batch size of $1$ and the weight decay of $10^{-4}$. 
We warm up the training for $25$K iterations, with the learning rate linearly increasing from $10^{-5}$ to $10^{-4}$. The learning rate then decreases to $10^{-5}$ with cosine annealing. We checkpoint the model every $50$K iterations and use the averaged model of the last $30$ checkpoints for evaluation.
Our model is trained on the hybrid data from 3DMatch, KITTI, nuScenes and ModelNet. For simplicity, we do not balance the samples from each dataset when constructing the training data.
The same augmentations as~\cite{qin2022geometric} are used in the experiments.
Following~\cite{yao2024pare}, we employ the random crop augmentation on 3DMatch. And we use the same augmentations as KITTI on nuScenes.
We do not use geometry-aware normalization during training for efficiency, and set the voxel size to $0.025$m for 3DMatch/ModelNet, and $0.3$m for KITTI and nuScenes.
To compensate, we adopt extra random scaling augmentation ranging from $[0.8, 1.2]$.

\subsection{Benchmarks}

\para{3DMatch.}
3DMatch~\cite{zeng20173dmatch} is a widely used indoor point cloud registration dataset, which consists of $62$ scenes captured by RGB-D sensors. We follow the standard split of $46$ scenes for training, $8$ for validation and $8$ for testing. We use the same approach as~\cite{huang2021predator} to generate point cloud pairs. During testing, the overlap ratios of the point cloud pairs in 3DMatch are above $30\%$, and those in 3DLoMatch are between $10\%$ and $30\%$.
There are $20641$ pairs for training, $1331$ for validation, $1623$ and $1781$ for testing on 3DMatch and 3DLoMatch, respectively.

\para{ScanNet++.}
ScanNet++~\cite{yeshwanth2023scannet++} is a large-scale 3D reconstruction dataset. We use the depth images captured using an iPhone 13 Pro. We use the same approach as~\cite{seo2025buffer} to process the dataset, resulting in $2050$ point cloud pairs for testing.

\para{TIERS.}
TIERS~\cite{li2022multi} is multi-modal SLAM dataset captured by various types of LiDAR scanners. Following~\cite{seo2025buffer}, we use the point clouds captured by Velodyne VLP-16, Ouster OS1-64, and Ouster OS0-128 from the sequences \texttt{Indoor06}-\texttt{Indoor11} (except \texttt{Indoor07}) sequences. We select the point cloud pairs which are at most $2$m apart, obtaining $870$ testing pairs.

\para{KITTI.}
KITTI~\cite{geiger2012we} is an autonomous driving dataset collected by a $64$-beam LiDAR sensor. Following~\cite{huang2021predator,qin2022geometric,seo2025buffer}, we use sequences $0$-$5$ for training, $6$-$7$ for validation and $8$-$10$ for testing. We use the point cloud pairs which are at most $10$m apart and refine their poses with ICP following previous work.
There are $1358$ training pairs, $180$ validation pairs, and $555$ testing pairs, respectively.

\para{nuScenes.}
nuScenes~\cite{caesar2020nuscenes} is also an autonomous driving dataset captured by a $32$-beam LiDAR scanner. We use the same method as KITTI to generate the point cloud pairs, and obtain $20587$/$1271$/$1330$ pairs for training, validation and testing, respectively.

\para{ETH.}
ETH~\cite{pomerleau2012challenging} contains $4$ outdoor scenes captured by a Hokuyo UTM-30LX LiDAR scanner. We follow the settings in~\cite{ao2023buffer,seo2025buffer}, which contains $713$ point cloud pairs for testing.


\begin{table}[!t]
\small
\centering
\caption{Metric thresholds of all benchmarks.}
\label{table:thresholds}
\begin{tabular}{lcccc}
\toprule
 & IR(m) & RRE($^{\circ}$) & RTE(m) & RE(m) \\
\midrule
3DMatch~\cite{zeng20173dmatch} & 0.1 & - & - & 0.2 \\
3DLoMatch~\cite{zeng20173dmatch,huang2021predator} & 0.1 & - & - & 0.2 \\
ScanNet++~\cite{yeshwanth2023scannet++} & 0.1 & 5.0 & 0.3 & - \\
TIERS~\cite{li2022multi} & 0.8 & 5.0 & 0.6 & - \\
KITTI~\cite{geiger2012we} & 1.0 & 5.0 & 0.6 & - \\
nuScenes~\cite{caesar2020nuscenes} & 1.0 & 5.0 & 0.6 & - \\
ETH~\cite{pomerleau2012challenging} & 0.8 & 2.0 & 0.3 & - \\
WOD~\cite{sun2020scalability} & 1.0 & 5.0 & 0.6 & - \\
MIT~\cite{tian2023resilient} & 1.0 & 5.0 & 0.6 & - \\
Oxford~\cite{ramezani2020newer} & 1.0 & 5.0 & 0.6 & - \\
ModelNet~\cite{wu20153d} & 0.1 & 5.0 & 0.1 & - \\
ShapeNet~\cite{chang2015shapenet} & 0.1 & 5.0 & 0.1 & - \\
\bottomrule
\end{tabular}
\end{table}

\para{WOD.}
Waymo Open Dataset~\cite{sun2020scalability} is a large-scale autonomous driving dataset published by Waymo. We follow the settings in~\cite{liu2024extend} to select the point cloud pairs which are at most $10$m apart. At a result, there are $130$ testing pairs.

\para{MIT.}
MIT comes from the \texttt{10\_14\_acl\_jackal} sequence of the Kimera-Multi dataset~\cite{tian2023resilient}, which present a outdoor campus environment. The point clouds are captured by a Velodyne VLP 16 sensor, which are relatively sparse. Following~\cite{seo2025buffer}, we use the point cloud pairs which are at most $5$m apart, resulting in $230$ pairs for testing.

\para{Oxford.}
Oxford is extracted from the Newer College dataset~\cite{ramezani2020newer}, which consists of the \texttt{01}, \texttt{05}, and \texttt{07} sequences.
We follow~\cite{seo2025buffer} to generate $301$ testing point cloud pairs by choosing those which are at most $5$m apart.

\para{ModelNet.}
ModelNet contains man-made CAD models from $40$ categories. We use the first $20$ categories for training/validation, and the rest $20$ for testing. The $8$ symmetric categories are ignored. We follow the data augmentation in~\cite{qin2022geometric} to generate the point cloud pairs. 
As a result, we have $4194$ CAD models for training, $1002$ for validation, and $1146$ for testing.

\para{ShapeNet.}
ShapeNet is a large-scale CAD model dataset organized according to the WordNet taxonomy. We use the ShapeNetCore subset, which contains $55$ categories of common objects. We randomly sample $50$ CAD models from each categories, obtaining $2750$ models for testing. For each model, we uniformly sample $8192$ points to produce the point cloud. And we employ the same data augmentation as ModelNet to generate the testing pairs.

\subsection{Metrics}

Following common practice~\cite{huang2021predator,qin2022geometric,yao2024pare,seo2025buffer}, we use different methods to compute the \emph{Registration Recall} for different benchmarks.
For 3DMatch and 3DLoMatch, a point cloud pair is considered successfully registered if its registration error is below $0.2$m.
For other benchmarks, a registration is successful if its relative rotation error is below $\tau_{\text{rre}}$ and its relative translation error is below $\tau_{\text{rte}}$. We use stricter thresholds than BUFFER-X~\cite{seo2025buffer} to better evaluate the performance of different methods. The threshold values for each benchmark are shown in \cref{table:thresholds}.

\subsection{Additaional Results}


\begin{table}[!t]
\small
\centering
\caption{Computational time on 3DMatch. All results are measured in seconds.}
\label{table:runtime}
\begin{tabular}{lcccc}
\toprule
\multirow{2}{*}{Model} & \multirow{2}{*}{\makecell{Geometry-aware\\Normalization}} & \multirow{2}{*}{\makecell{Data\\Preparation}} & \multirow{2}{*}{\makecell{Model\\Inference}} & \multirow{2}{*}{Total} \\
 &  &  &  &  \\
\midrule
GeoTrans~\cite{qin2022geometric} & - & 0.246 & 0.062 & 0.308 \\
BUFFER~\cite{ao2023buffer} & - & 0.173 & 0.148 & 0.321 \\
BUFFER-X~\cite{seo2025buffer} & - & 0.133 & 0.752 & 0.885 \\
PARENet~\cite{yao2024pare} & - & 0.234 & 0.075 & 0.309 \\
RAM (\emph{ours}) & 0.212 & 0.258 & 0.055 & 0.525 \\
\bottomrule
\end{tabular}
\end{table}


\begin{table}[!t]
\small
\centering
\caption{Detailed computational time of \ours on 3DMatch. All results are measured in seconds.}
\label{table:runtime-details}
\begin{tabular}{cccccc}
\toprule
\multirow{2}{*}{\makecell{Geometry-aware\\Normalization}} & \multirow{2}{*}{\makecell{Data\\Preparation}} & \multirow{2}{*}{\makecell{Feature\\Extraction}} & \multirow{2}{*}{\makecell{Superpoint\\Matching}} & \multirow{2}{*}{\makecell{Point\\Matching}} & \multirow{2}{*}{\makecell{Local-to-Global\\Registration}} \\
 &  &  &  &  &  \\
\midrule
0.212 & 0.258 & 0.014 & 0.022 & 0.010 & 0.009 \\
\bottomrule
\end{tabular}
\end{table}

\para{Computational efficiency.}
We compare the computational time for GeoTrans~\cite{qin2022geometric}, BUFFER~\cite{ao2023buffer}, BUFFER-X~\cite{seo2025buffer}, PARENet~\cite{yao2024pare}, and \ours in \cref{table:runtime}, averaged over all test cases in 3DMatch. \ours has the fastest model-inference time among the compared methods, supporting the computational advantage of \egtr. And the detailed computational time of \ours is shown in \cref{table:runtime-details}. The first two stages are executed on the CPU, while the remaining stages run on the GPU. Geometry-aware normalization takes a comparable amount of time to data preparation (\ie, building hierarchical computation graphs for point clouds). In particular, we observe that most of the runtime of geometry-aware normalization is spent on constructing KD-trees and querying neighbors on the raw point clouds. In our current implementation, the KD-trees for the raw point clouds are rebuilt at every iteration, Caching or reusing the neighborhood structures should reduce this overhead. This stage can also be accelerated by pre-voxelization with a small voxel size (\eg, $0.01$m). In practice, all stages can be pipelined, yielding an inference speed of around $4$ fps.

\para{Volume of training data.}
As the hybrid data contains more training samples (approximately $20$K pairs from 3DMatch, $1.3$K pairs from KITTI, $20$K pairs from nuScenes, and $4$K pairs from ModelNet), we then extend the training of the indoor-only \ours to $120$ epochs, which results in approximately the same total number of sample exposures as the hybrid model while retaining the same architecture and optimization settings. As shown in \cref{table:results-volume}, even after matching the total training exposure, the hybrid model consistently outperforms the indoor-only model on both 3DMatch and 3DLoMatch. These results indicate that the gains mainly come from the cross-domain training, rather than solely the increase of total training samples.


\begin{table}[!t]
\small
\centering
\caption{
Results on 3DMatch and 3DLoMatch with extended training samples.
\textbf{Boldfaced} numbers highlight the best.
}
\label{table:results-volume}
\begin{tabular}{lccccc}
\toprule
\multirow{2}{*}{Data} & \multirow{2}{*}{\# Epochs} & \multicolumn{2}{c}{3DMatch} & \multicolumn{2}{c}{3DLoMatch} \\
\cmidrule(r){3-6}
 & & IR(\%) & RR(\%) & IR(\%) & RR(\%) \\
\midrule
indoor & 60 & 75.0 & 92.4 & 47.7 & 77.8 \\
indoor & 120 & 76.9 & 92.8 & 50.3 & 79.2 \\
hybrid & 60 & \textbf{79.7} & \textbf{93.3} & \textbf{53.7} & \textbf{81.1} \\
\bottomrule
\end{tabular}
\end{table}

\para{Rotation invariance.}
We further evaluate the robustness of \ours to large rotations on 3DMatch and 3DLoMatch.
Following~\cite{yu2023rotation,yao2024pare}, we apply random rotations sampled from $\mathcal{U}(-\pi, \pi)$.
We denote the original setting as standard and the large-rotation setting as rotated. We report RR by directly computing registration RMSE instead of estimating from covariance matrices~\cite{zeng20173dmatch}, so the RR values are slightly different those reported in the main paper. For a fair comparison, we evaluate both the standard and rotated settings using the same protocol.


\begin{table}[!t]
\small
\centering
\caption{
Results on 3DMatch with large rotations.
}
\label{table:results-rotated}
\begin{tabular}{lcccccccc}
\toprule
\multirow{2}{*}{Model} & \multicolumn{2}{c}{3DMatch} & \multicolumn{2}{c}{3DLoMatch} \\
\cmidrule(r){2-5}
 & IR(\%) & RR(\%) & IR(\%) & RR(\%) \\
\midrule
PARENet~\cite{yao2024pare} (standard) & 73.6 & 97.0 & 44.3 & 80.5 \\
PARENet~\cite{yao2024pare} (rotated) & 73.7 (\textcolor{green}{0.1$\uparrow$}) & 96.7 (\textcolor{red}{0.3$\downarrow$}) & 44.8 (\textcolor{green}{0.5$\uparrow$}) & 81.1 (\textcolor{green}{0.6$\uparrow$}) \\
\midrule
\ours (\emph{ours}, standard) & 79.7 & 96.2 & 53.7 & 82.4 \\
\ours (\emph{ours}, rotated) & 78.9 (\textcolor{red}{0.8$\downarrow$}) & 95.5 (\textcolor{red}{0.7$\downarrow$}) & 52.7 (\textcolor{red}{1.0$\downarrow$}) & 79.7 (\textcolor{red}{2.7$\downarrow$}) \\
\bottomrule
\end{tabular}
\end{table}

As shown in \cref{table:results-rotated}, PARENet remains stable under large rotations because of its rotation-equivariant design. \ours experiences larger, but still limited, degradation: \emph{Inlier Ratio} decreases by $0.8$ and $1.0$ points on 3DMatch and 3DLoMatch, respectively, while \emph{Registration Recall} decreases by $0.7$ and $2.7$ points. Importantly, \ours retains strong absolute performance under large rotations and continues to outperform PARENet in \emph{Inlier Ratio} on both benchmarks. We attribute this robustness to two factors. First, we apply rotation augmentation during training, which improves the efficacy of the KPConv backbone under large rotations. Second, \ours encodes geometric consistency during both superpoint matching and point matching, further mitigating the impact of the rotation-variant backbone.

\para{Cross-source registration.}
We also evaluate the generalization ability of \ours to cross-source registration on the 3DCSR benchmark~\cite{huang2103comprehensive}, following the preprocessing protocols of~\cite{zhao2024vrhcf,zhao2025cross}. As in \cref{table:results-3dcsr}, although \ours is not trained specifically for cross-source registration, it substantially outperforms GeoTransformer on both subsets, especially on the more challenging LiDAR subset. These results indicate that the cross-domain training strategy also provides meaningful generalization benefits under cross-source shifts.


\begin{table}[!t]
\small
\centering
\caption{
Results on 3DCSR benchmark.
\textbf{Boldfaced} numbers highlight the best and the second best are \underline{underlined}.
}
\label{table:results-3dcsr}
\begin{tabular}{lcccccccc}
\toprule
\multirow{2}{*}{Model} & \multicolumn{2}{c}{SfM} & \multicolumn{2}{c}{LiDAR} \\
\cmidrule(r){2-5}
 & IR(\%) & RR(\%) & IR(\%) & RR(\%) \\
\midrule
GeoTrans~\cite{qin2022geometric} & 40.2 & 84.4 & 3.4 & 9.1 \\
Cross-PCR~\cite{zhao2025cross} & \textbf{89.4} & \textbf{93.8} & \textbf{35.5} & \textbf{66.7} \\
\midrule
\ours (\emph{ours}) & \underline{44.9} & \underline{90.6} & \underline{13.3} & \underline{42.9} \\
\bottomrule
\end{tabular}
\end{table}

However, \ours remains behind Cross-PCR, which is specifically designed for cross-source registration. We attribute part of this gap to the sensitivity of our current encoder to severe density and sampling discrepancies between heterogeneous sources. Extending \ours with cross-source registration models may further improve its robustness to heterogeneous modalities. However, this is beyond the scope of this work and we will leave it as future work.

\begin{figure*}[t]
  \centering
  \begin{overpic}[width=\linewidth]{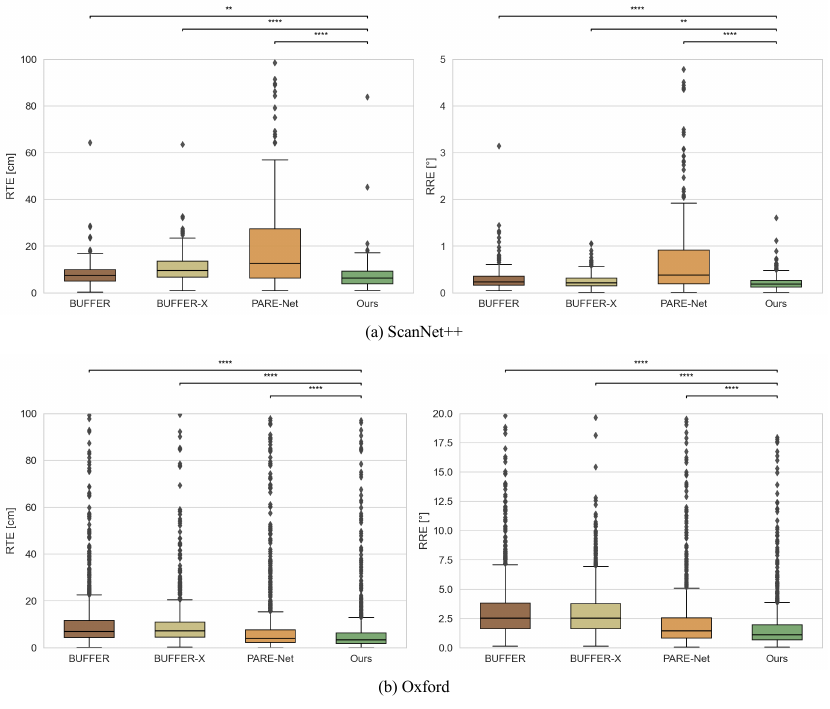}
  \end{overpic}
  \caption{Statistical tests on ScanNet++ and Oxford benchmarks.}
  \label{fig:significance}
\end{figure*}

\para{Statistical test.}
Following~\cite{seo2025buffer}, we conduct statistical comparisons on ScanNet++ and Oxford against BUFFER, BUFFER-X, and PARENet. For ScanNet++, we use the official baseline checkpoints trained on 3DMatch. For Oxford, we use the corresponding checkpoints trained on KITTI. As shown in \cref{fig:significance}, \ours achieves the lowest median \emph{RTE} and \emph{RRE} with the smallest variance. And the statistical tests confirm that the improvements over all baselines are highly significant.

\subsection{Discussions}

\para{Trade-off between generalization and per-domain accuracy.}
To examine the influence of cross-domain training on per-domain accuracy, \cref{table:results-ablation}~(a) of the main paper compares the hybrid model with domain-specific models using the same architecture and optimization protocol. The hybrid model consistently outperforms the corresponding domain-specific models across the evaluated benchmarks. These results indicate that, under our training design, cross-domain training does not introduce a systematic trade-off between generalization and per-domain accuracy.

At the same time, domain-relevant training data remain important for performance within each domain. As shown in \cref{table:results-outdoor}, the complete hybrid model achieves the best overall performance, showing that cross-domain diversity is beneficial rather than detrimental. However, removing nuScenes (marked by $^{\diamond}$) leads to a performance decrease, particularly on MIT, while the outdoor-only model (marked by $^{star}$) partially recovers this loss, improving \emph{Inlier Ratio} from $62.5$ to $72.3$ and RR from $97.4$ to $98.7$. It also matches the hybrid model without nuScenes on WOD.

These results suggest that cross-domain and domain-relevant data play complementary roles: cross-domain data expose the model to more diverse geometric patterns and improve broad generalization, whereas data from the target domain provide important domain-specific structures and sampling characteristics.

\para{Effect of LiDAR sparsity on low-overlap registration.}
We believe that the improvement on 3DLoMatch primarily arises from the greater geometric diversity introduced by cross-domain training, rather than simply from a larger number of training-sample exposures. As shown in \cref{table:results-volume}, extending the training of the indoor-only model to approximately match the total sample exposure of the hybrid model does not close the performance gap on 3DLoMatch. The hybrid model still improves \emph{Inlier Ratio} from $50.3$ to $53.7$ and \emph{Registration Recall} from $79.2$ to $81.1$.

Although both sparsity and low overlap lead to incomplete observations, they actually impose distinct difficulties on registration: sparsity weakens the local geometric cues within each point cloud, whereas low overlap reduces the shared structure between the pair and increases correspondence ambiguity in non-overlapping regions. Moreover, most outdoor training pairs still have relatively high overlap. Therefore, introducing sparse LiDAR data does not directly expose the model to the same challenge as low-overlap registration.

Instead, the sparse and incomplete observations in LiDAR data enrich the geometric and sampling patterns during training. This diversity may enhances feature discriminativeness and improves the model's robustness to partial observations, therefore benefiting low-overlap registration indirectly.

\subsection{Broader Impacts}

This work advances robust cross-domain 3D point cloud registration, enabling accurate spatial alignment across diverse sensors without extensive per-domain fine-tuning or re-training. It can act as a LiDAR odometry component in 3D vision systems and benefit broad applications, such as autonomous navigation, augmented and virtual reality, digital twins, construction monitoring, cultural heritage preservation, etc.
Another potential impact is that this work unveils the efficacy of cross-domain training in point cloud processing tasks. The geometry-aware normalization enables processing point clouds from diverse domains in a unified model, which could act as a fundamental component towards point cloud foundation models.

\end{document}